\documentclass[10pt,twocolumn,letterpaper]{article}

\usepackage{alex}
\usepackage{cvpr}
\usepackage{times}
\usepackage{epsfig}
\usepackage{graphicx}
\usepackage{amsmath}
\usepackage{amssymb}
\usepackage{tikz,pgfplots}
\usepackage{cite}
\usepackage{tango}
\usepackage{tabulary,multirow,overpic,xcolor}
\usepackage[title]{appendix}
\usepackage[caption=false]{subfig}
\def\x{\times}

\newcommand{\stm}{\mathrm{STO}}
\newcommand{\ltm}{\mathrm{FBO}}
\newcommand{\ltmnl}{\ltm_{\textrm{NL}}}
\newcommand{\ltmpool}{\ltm_{\textrm{pool}}}
\newcommand{\ltmavg}{\ltm_{\textrm{Avg}}}
\newcommand{\ltmmax}{\ltm_{\textrm{Max}}}

\newcommand{\nlmod}{\mathrm{NL'}}

\usepgfplotslibrary{fillbetween}
\pgfplotsset{compat = 1.3,
          legend style={font=\scriptsize},
          legend cell align={left},
          legend style={cells={align=left}, draw=black!20},
          grid=both,grid style={dotted},tick style={draw=none},
          enlarge x limits=false,
          enlarge y limits=false,
          axis line style={draw=black!100},
          }

\pgfplotsset{ every non boxed x axis/.append style={x axis line style=-},
              every non boxed y axis/.append style={y axis line style=-}}
\pgfplotscreateplotcyclelist{mystylelist}{%
densely dashdotted\\
dotted\\
densely dashed\\
dashed\\
solid\\
dashdotted\\
densely dotted\\
densely dash dot dot\\
dotted\\
dash dot dot\\
}
\pgfplotscreateplotcyclelist{mycolorlist}{%
chameleon3\\
skyblue2\\
scarletred3\\
orange2\\
plum1\\
aluminium6\\
skyblue3\\
chocolate2\\
maroon\\
butter3\\
}
\pgfplotscreateplotcyclelist{mycolormarklist}{%
chameleon3,mark=triangle*,\\
skyblue2,mark=square*,mark options={solid}\\
scarletred3,mark=diamond*,mark options={solid}\\
aluminium5,mark=otimes*,mark options={solid}\\
plum1,mark=triangle*,mark options={solid}\\
orange1,mark=*,mark options={solid}\\
skyblue3\\
chocolate2\\
maroon\\
butter3\\
}

\newcolumntype{x}[1]{>{\centering\arraybackslash}p{#1pt}}

\newlength\savewidth\newcommand\shline{\noalign{\global\savewidth\arrayrulewidth
  \global\arrayrulewidth 1pt}\hline\noalign{\global\arrayrulewidth\savewidth}}
\newcommand{\tablestyle}[2]{\setlength{\tabcolsep}{#1}\renewcommand{\arraystretch}{#2}\centering\footnotesize}
\makeatletter\renewcommand\paragraph{\@startsection{paragraph}{4}{\z@}
  {.5em \@plus1ex \@minus.2ex}{-.5em}{\normalfont\normalsize\bfseries}}\makeatother
\definecolor{demphcolor}{RGB}{100,100,100}
\newcommand{\demph}[1]{\textcolor{demphcolor}{#1}}

\definecolor{darkgreen}{rgb}{0.0, 0.5, 0.0}
\usepackage[pagebackref=true,breaklinks=true,letterpaper=true,colorlinks,bookmarks=false,citecolor=darkgreen]{hyperref}

\def\cvprPaperID{****}\cvprfinalcopy

\begin{document}

\title{Long-Term Feature Banks for Detailed Video Understanding}

\author{%
Chao-Yuan Wu$^{1,2}$ \quad\quad\quad Christoph Feichtenhofer$^2$ \quad\quad\quad Haoqi Fan$^2$ \\
\ \ \ \quad Kaiming He$^2$ \quad\quad\quad\quad\quad Philipp Kr\"ahenb\"uhl$^1$ \ \quad\quad\quad Ross Girshick$^2$ \vspace{.5em} \\
$^1$The University of Texas at Austin  \quad $^2$Facebook AI Research (FAIR) \vspace{-.5em}
}

\maketitle

%%%%%%%%%%%%%%%%%%%%%%%%%%%%%%%%%%%%%%%%%%%%%%%%%%%%%%%%%%%%%%%%%%%%%%%%%%%%%%%%%%%%%%%%%%%%%%%%%%%
\begin{abstract}
\vspace{-1.5mm}
To understand the world, we humans constantly need to relate the present to the past, and put events in context. 
In this paper, we enable existing video models to do the same.
We propose a \emph{long-term feature bank}---supportive information extracted over the entire span of a video---to augment
state-of-the-art video models that otherwise would only view short clips of 2-5 seconds.
Our experiments demonstrate that augmenting 3D convolutional networks with a long-term feature bank yields state-of-the-art results on three challenging video datasets: AVA, EPIC-Kitchens, and Charades.
Code is available online.\footnote{\url{https://github.com/facebookresearch/video-long-term-feature-banks}}
\end{abstract}
\vspace{-2.9mm}

%%%%%%%%%%%%%%%%%%%%%%%%%%%%%%%%%%%%%%%%%%%%%%%%%%%%%%%%%%%%%%%%%%%%%%%%%%%%%%%%%%%%%%%%%%%%%%%%%%%
\section{Introduction}
What is required to understand a movie? Many aspects of human intelligence, for sure, but \emph{memory} is particularly important. As a film unfolds there's a constant need to relate whatever is happening in the present to what happened in the past. Without the ability to use the past to understand the present, we, as human observers, would not understand what we are watching.

\begin{figure}[t]
\centering
\includegraphics[width=1.0\linewidth]{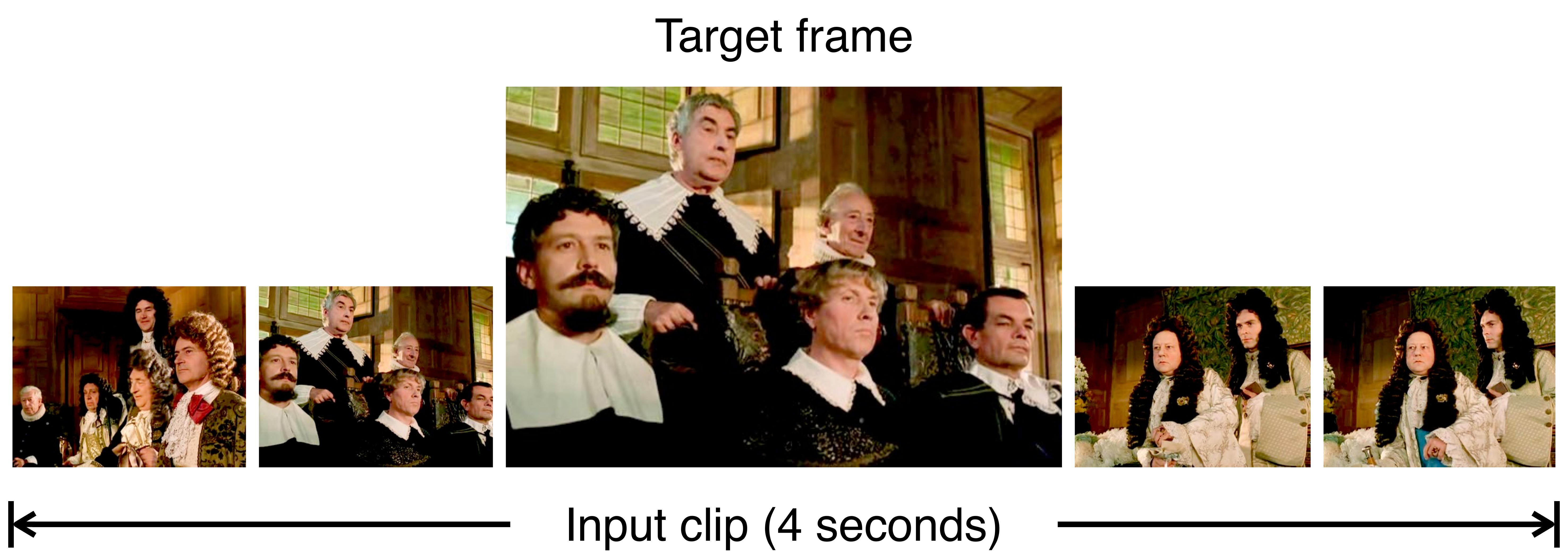}
\caption{What are these people doing? Current 3D CNN video models operate on short clips spanning only $\sim$4 seconds. 
Without observing longer-term context, recognition is difficult. (Video from the AVA dataset~\cite{ava}; see next page for the answer.)}
\label{fig:overview1}
\end{figure}

\begin{figure*}[t]
\centering
\includegraphics[width=1.0\linewidth]{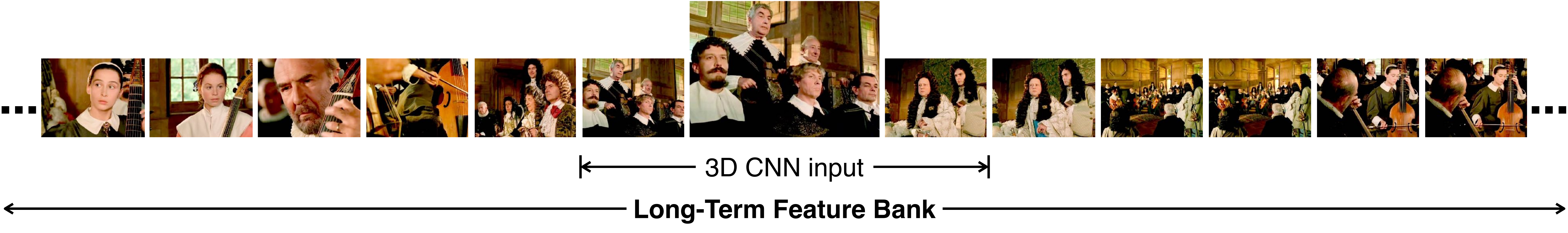}
\caption{The action becomes clear when relating the target frame to the long-range context. 
Our \textbf{long-term feature bank} provides long-term supportive information that enables video models to better understand the present.
(AVA ground-truth label: `listening to music'.)}
\label{fig:overview2}
\vspace{-2mm}
\end{figure*}

In this paper, we propose the idea of a \emph{long-term feature bank} that stores a rich, time-indexed representation of the entire movie. Intuitively, the long-term feature bank stores features that encode information about past and (if available) future scenes, objects, and actions. This information provides a supportive context that allows a video model, such as a 3D convolutional network, to better infer what is happening in the present (see~\figref{overview1},~\ref{fig:overview2}, and~\ref{fig:viz}).

We expect the long-term feature bank to improve state-of-the-art video models because most make predictions based only on information from a \emph{short} video clip, typically 2-5 seconds~\cite{tran2015learning,i3d,xie2017rethinking,varol2018long,tran2017convnet,qiu2017learning,strg,nonlocal}. The reason for this short-term view is simple: benchmark advances have resulted from training \emph{end-to-end} networks that use some form of 3D convolution, and these 3D convolutions require dense sampling in time to work effectively. Therefore, to fit in GPU memory the video inputs must be short.

%##################################################################################################
%##################################################################################################

%##################################################################################################
%##################################################################################################

The long-term feature bank is inspired by works that leverage long-range temporal information by using pre-computed visual features~\cite{yue2015beyond,li2017temporal,miech2017learnable,tang2018non}.
However, these approaches extract features from isolated frames using an ImageNet pre-trained network, and then use the features as input into a trained pooling or recurrent network. Thus the \emph{same} features represent both the present and the long-term context. In contrast, we propose to \emph{decouple} the two: the long-term feature bank is an auxiliary component that \emph{augments} a standard video model, such as a state-of-the-art 3D CNN\@.
This design enables the long-term feature bank to store flexible supporting information, such as object detection features, that differ from what the 3D CNN computes.

Integrating the long-term feature bank with 3D CNNs is straightforward. We show that a variety of mechanisms are possible, including an attentional mechanism that relates information about the present (from the 3D CNN) to long-range information stored in the long-term feature bank. We illustrate its application to different tasks with different output requirements: we show results on datasets that require object-level as well as frame- or video-level predictions.

Finally, we conduct extensive experiments demonstrating that augmenting 3D CNNs with a long-term feature bank yields state-of-the-art results on three challenging video datasets: AVA spatio-temporal action localization~\cite{ava}, EPIC-Kitchens verb, noun, and action classification~\cite{epic}, and Charades video classification~\cite{charades}. Our ablation study establishes that the improvements on these tasks arise from the integration of long-term information.

%%%%%%%%%%%%%%%%%%%%%%%%%%%%%%%%%%%%%%%%%%%%%%%%%%%%%%%%%%%%%%%%%%%%%%%%%%%%%%%%%%%%%%%%%%%%%%%%%%%
\section{Related Work}

\paragraph{Deep networks} are the dominant approach for video understanding~\cite{karpathy2014large,simonyan2014two,tsn,tran2015learning,i3d,xie2017rethinking,varol2018long,tran2017convnet,qiu2017learning,nonlocal}.
This includes the highly successful two-stream networks~\cite{karpathy2014large,simonyan2014two,tsn} and 3D convolutional networks~\cite{tran2015learning,i3d,xie2017rethinking,varol2018long,tran2017convnet,qiu2017learning,nonlocal}. 
In this paper, we use 3D CNNs, but the long-term feature bank 
can be integrated with other families of video models as well. 

\paragraph{Temporal and relationship models} include RNNs that model the evolution of video frames~\cite{donahue2015long,yue2015beyond,sun2017lattice,li2018videolstm,li2018recurrent} and 
multilayer perceptrons that model ordered frame features~\cite{zhou2017temporalrelation}.
To model finer-grained interactions, 
a growing line of work leverages pre-computed object proposals~\cite{strg} or detections~\cite{baradel2018object,ma2017attend},
and models their co-occurrence~\cite{sun2018actor,strg,ma2017attend}, temporal order~\cite{baradel2018object}, or spatial arrangement~\cite{strg}
within a short clip. 

\paragraph{Long-term video understanding} with modern CNNs is less explored, in part due to GPU memory constraints.
One strategy to overcome these constraints is to
use pre-computed features without end-to-end training~\cite{yue2015beyond,li2017temporal,miech2017learnable,tang2018non}.
These methods do not optimize features for a target task, and thus are likely suboptimal. 
Another strategy is to use aggressive subsampling~\cite{tsn,zhou2017temporalrelation} or large striding~\cite{feichtenhofer2016spatiotemporal}. 
TSN~\cite{tsn} samples 3-7 frames per video. 
ST-ResNet~\cite{feichtenhofer2016spatiotemporal} uses a temporal stride of 15.
To our knowledge, our approach is the first
that enjoys the best of three worlds: 
end-to-end learning for strong short-term features with
dense sampling \emph{and} decoupled, flexible long-term modeling.  

\paragraph{Spatio-temporal action localization} is an active research area~\cite{ava,gkioxari2015finding,singh2017online,peng2016multi,hou2017tube,weinzaepfel2015learning}.
Most recent approaches extend object detection frameworks~\cite{girshick2015fast,Ren2015a} to
first propose tubelets/boxes in a short clip/frame, and then
classify the tubelets/boxes into action classes~\cite{ava,kalogeiton2017action,saha2017amtnet,peng2016multi,saha2017amtnet,hou2017tube}.
The detected tubelets/boxes can then be optionally linked to form full action tubes~\cite{gkioxari2015finding,peng2016multi,saha2017amtnet,kalogeiton2017action,singh2017online,hou2017tube}.
In contrast to our method, these methods find actions within each frame or clip independently without exploiting long-term context. 

\paragraph{Information `bank'} representations, such as \emph{object bank}~\cite{li2010object}, \emph{detection bank}~\cite{althoff2012detection}, and \emph{memory networks}~\cite{sukhbaatar2015end} have been used as image-level representations, for video indexing and retrieval, and for modeling information in text corpora. We draw inspiration from these approaches and develop methodologies for detailed video understanding tasks.

%##################################################################################################
\begin{figure*}[thb]\vspace{-3mm}
\centering
\subfloat[\textbf{3D CNN}\quad\quad\quad\label{fig:model:baseline}]{%
\includegraphics[height=3.8cm,page=1]{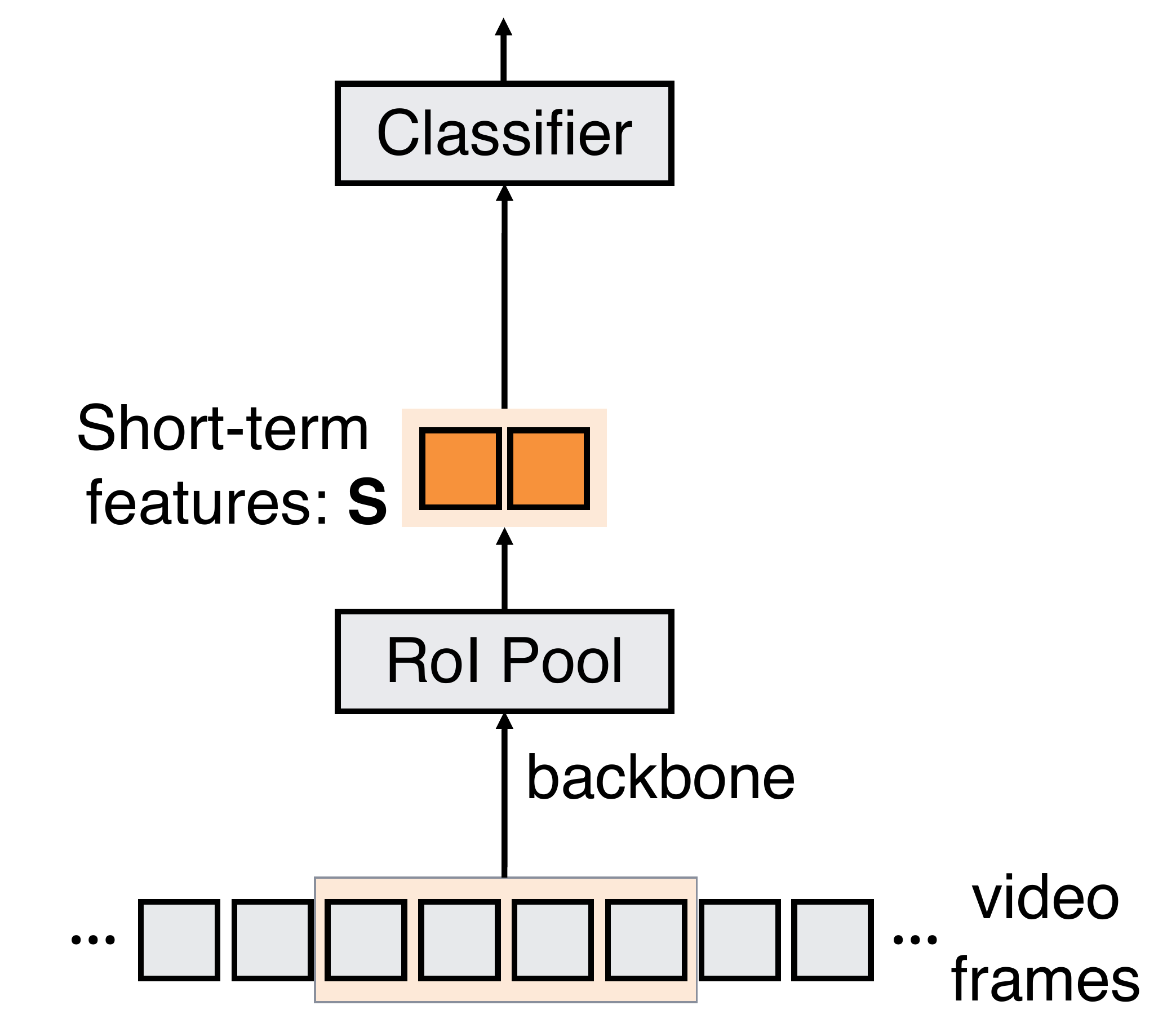}}\hspace{2cm}
\subfloat[\textbf{3D CNN with a Long-Term Feature Bank (Ours)}\label{fig:model:ltm}]{%
\includegraphics[height=3.8cm,page=1]{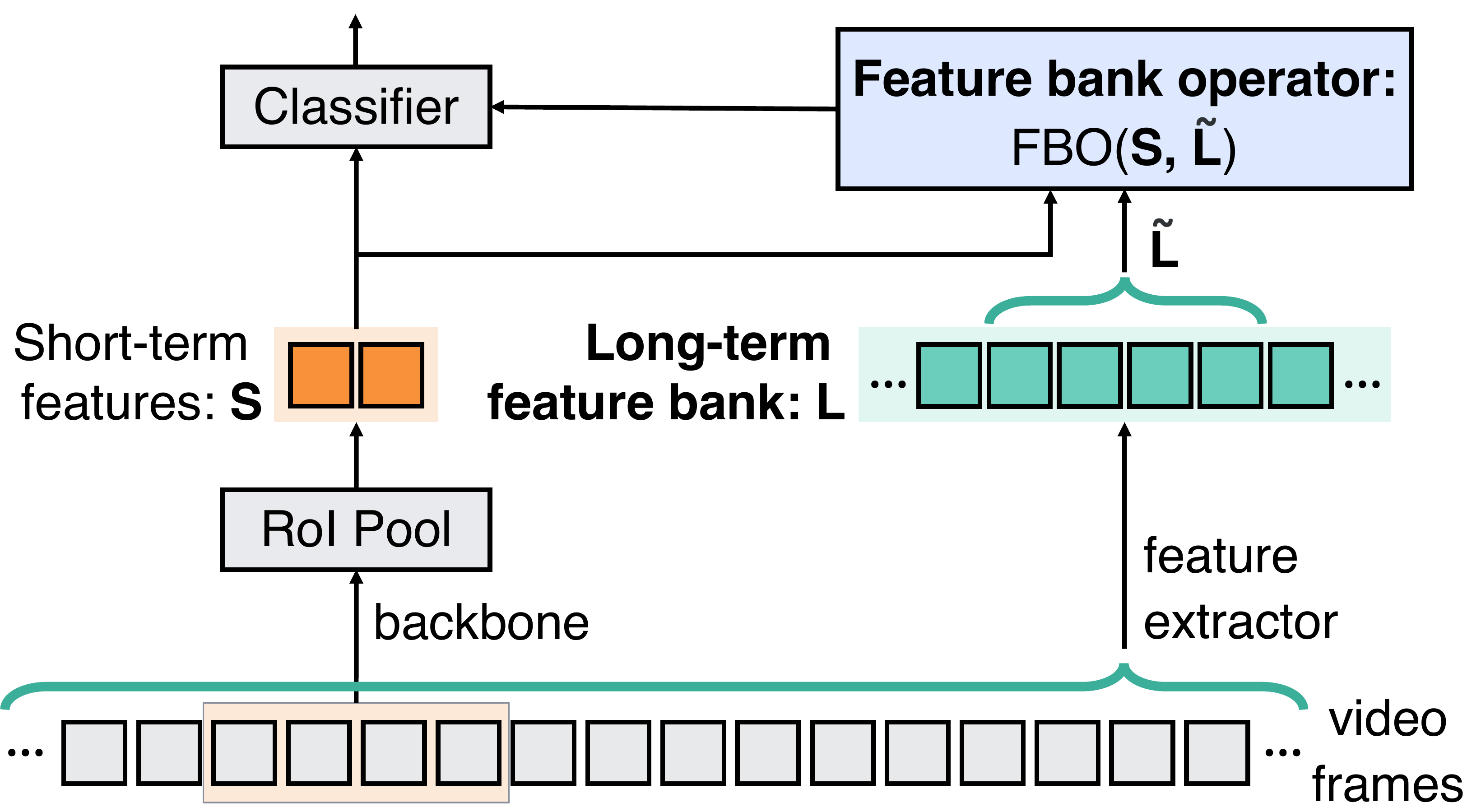}}
\vspace{1.5mm}
\caption{We contrast our model with standard methods. 
  (a) \textbf{3D CNN:} vanilla 3D CNNs process a \emph{short} clip from a video (\eg, 2-5 seconds) and use pooling
  to obtain a representation of the clip (\eg,~\cite{tran2015learning,i3d,nonlocal}). 
  (b) \textbf{Long-Term Feature Bank (Ours):} We extend the vanilla 3D CNN with 
  a long-term feature bank $L$ and a feature bank operator $\ltm(S, \tilde{L})$ that computes interactions between the short-term and long-term features. Our model is able to integrate information over a long temporal support, lasting minutes or even the whole video. }
\vspace{-3mm}
\end{figure*}
%##################################################################################################

%%%%%%%%%%%%%%%%%%%%%%%%%%%%%%%%%%%%%%%%%%%%%%%%%%%%%%%%%%%%%%%%%%%%%%%%%%%%%%%%%%%%%%%%%%%%%%%%%%%
\section{Long-Term Feature Bank Models}
\label{sec:method}
For computer vision models to make accurate predictions on long, complex videos they will certainly require the ability to relate what is happening in the present to events that are distant in time. With this motivation in mind, we propose a model with a \emph{long-term feature bank} to explicitly enable these interactions.

\subsection{Method Overview}
We describe how our method can be used for the task of \emph{spatio-temporal action localization}, where the goal is to detect all actors in a video and classify their actions. Most state-of-the-art methods~\cite{ava,girdhar2018better,sun2018actor} combine a `backbone' 3D CNN (\eg, C3D~\cite{tran2015learning}, I3D~\cite{i3d}) with a region-based person detector (\eg, Fast/Faster R-CNN~\cite{girshick2015fast,Ren2015a}). To process a video, it is split into \emph{short} clips of 2-5 seconds, which are \emph{independently} forwarded through the 3D CNN to compute a feature map, which is then used with region proposals and region of interest (RoI) pooling to compute RoI features for each candidate actor~\cite{ava,girdhar2018better}. This approach, which captures only short-term information, is depicted in \figref{model:baseline}.
 
The central idea in our method is to extend this approach with two new concepts: (1) a \textbf{long-term feature bank} that intuitively acts as a `memory' of what happened during the entire video---we compute this as RoI features from detections at regularly sampled time steps; and (2) a \textbf{feature bank operator} (FBO) that computes interactions between the short-term RoI features (describing what actors are doing now) and the long-term features. The interactions may be computed through an attentional mechanism, such as a non-local block~\cite{nonlocal}, or by feature pooling and concatenation. Our model is summarized in \figref{model:ltm}. We introduce these concepts in detail next.

\subsection{Long-Term Feature Bank}
The goal of the long-term feature bank, $L$, is to provide relevant contextual information to aid recognition at the current time step.
For the task of spatio-temporal action localization, we run a person detector over the entire video to generate a set of detections for each frame. In parallel, we run a standard clip-based 3D CNN, such as C3D~\cite{tran2015learning} or I3D~\cite{i3d}, over the video at regularly spaced intervals, such as every one second.
We then use RoI pooling to extract features for all person detections at each time-step processed by the 3D CNN\@.
Formally, $L = \sbr{L_0, L_1, \dots, L_{T-1}}$ is a time-indexed list of features for video time steps $0, \dots, T-1$, where $L_t\in \RR^{N_t \times d}$ is the matrix of $N_t$ $d$-dimensional RoI features at time $t$. 
Intuitively, $L$ provides information about when and what all actors are doing \emph{in the whole video} and it can be efficiently computed in a single pass over the video by the detector and 3D CNN\@.

\subsection{Feature Bank Operator}
Our model references information from the long-term features $L$ via a \emph{feature bank operator}, $\ltm(S_t,\tilde{L}_t)$. 
The feature bank operator accepts inputs $S_t$ and $\tilde{L}_t$, where $S_t$ is the short-term RoI pooled feature and $\tilde{L}_t$ is $\sbr{L_{t-w}, \dots, L_{t+w}}$, a subset of $L$ centered at the current clip at $t$ within `window' size $2w + 1$, stacked into a matrix $\tilde{L}_t \in \RR^{N \times d}$, where $N = \sum_{t'=t-w}^{t+w} N_{t'}$. 
We treat the window size $2w + 1$ as a hyperparameter that we cross-validate in our experiments.
The output is then channel-wise concatenated with $S_t$ and used as input into a linear classifier.

Intuitively, the feature bank operator computes an updated version of the pooled short-term features $S_t$ by relating them to the long-term features. The implementation of $\ltm$ is flexible. Variants of attentional mechanisms are an obvious choice and we will consider multiple instantiations in our experiments.

\paragraph{Batch \vs Casual.} Thus far we have assumed a \emph{batch} setting in which the entire video is available for processing. Our model is also applicable to online, \emph{casual} settings.
In this case, $\tilde{L}_t$ contains only past information of window size $2w+1$; we consider both batch and causal modes of operation in our experiments. 

\subsection{Implementation Details}

\paragraph{Backbone.} We use a standard 3D CNN architecture from recent video classification work. The model is a ResNet-50~\cite{He2016} that is pre-trained on ImageNet~\cite{Russakovsky2015} and `inflated' into a network with 3D convolutions (over space and time) using the I3D technique~\cite{i3d}. The network structure is modified to include non-local operations~\cite{nonlocal}. 
After inflating the network from 2D to 3D, we pre-train it for video classification on the Kinetics-400 dataset~\cite{i3d}. The model achieves 74.9\% (91.6\%) top-1 (top-5) accuracy on the Kinetics-400~\cite{i3d} validation set. Finally, we remove the temporal striding for conv$_1$ and pool$_1$ following~\cite{strg},
and remove the Kinetics-specific classification layers to yield the backbone model. 
The exact model specification is given in the Appendix.
The resulting network accepts an input of shape $32 \times H \times W \times 3$, representing 32 RGB frames with spatial size $H \times W$, and outputs features with shape $16 \times H/16 \times W/16 \times 2048$.
The same architecture is used to compute short-term features $S$ and long-term features $L$. Parameters are not shared between these two models unless otherwise noted.

\paragraph{RoI Pooling.} 
We first average pool the video backbone features over the time axis. We then use RoIAlign~\cite{He2017} with a spatial output of $7 \x 7$, followed by spatial max pooling, to yield a single 2048 dimensional feature vector for the RoI. 
This corresponds to using a temporally straight tube~\cite{ava}.

\paragraph{Feature Bank Operator Instantiations.} The feature bank operator can be implemented in a variety of ways. We experiment with the following choices; others are possible.

\begin{figure}[t]
\centering
\includegraphics[width=0.66\linewidth, page=1]{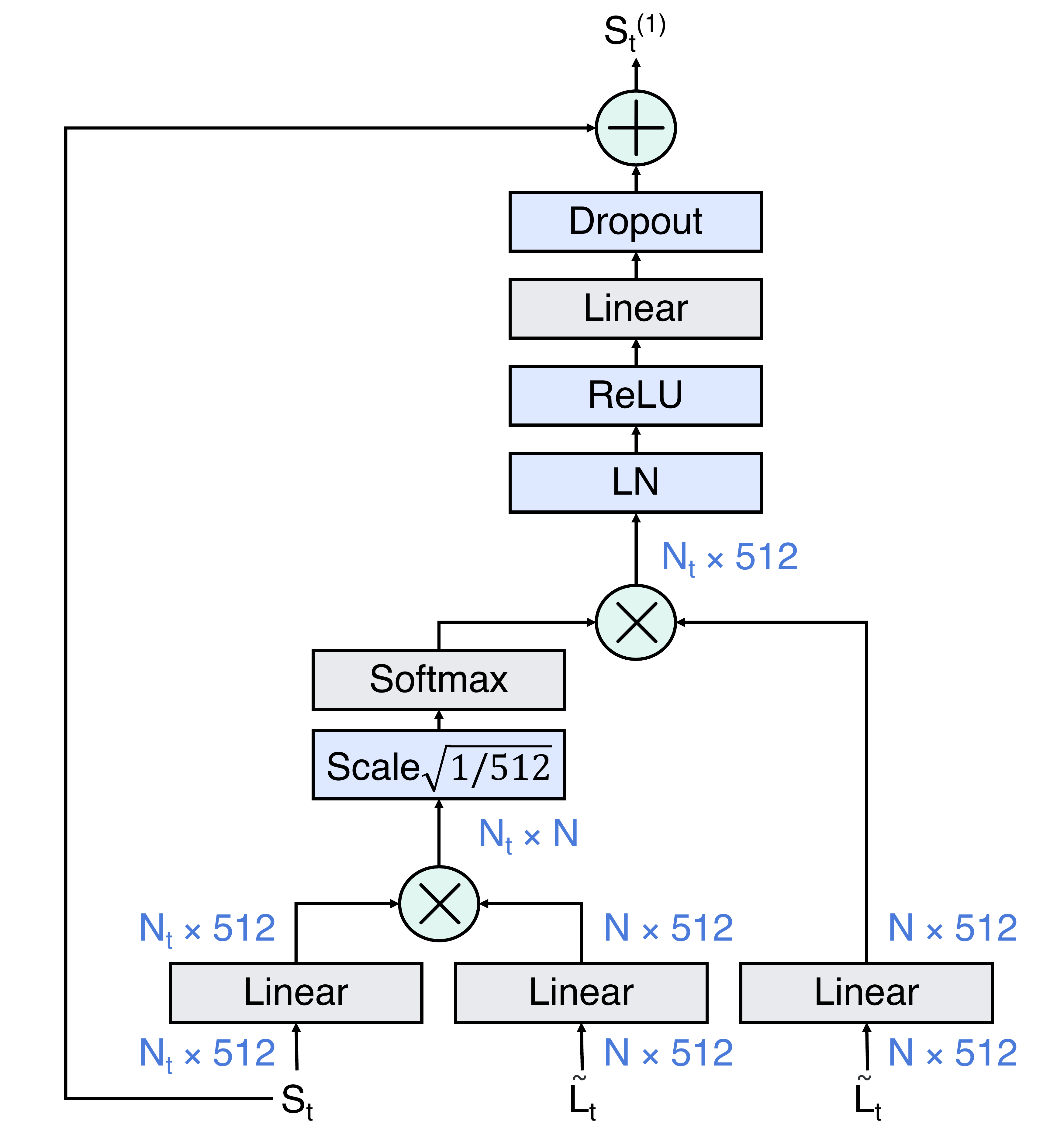}
\caption{\textbf{Our modified non-local block design.} Here we plot the first layer $S_t^{(1)} = \nlmod_{\theta_1}(S_t, \tilde{L}_t)$ as an example. 
  `$\otimes$' denotes matrix multiplication, and `$\oplus$' denotes element-wise sum. }
\label{fig:nl_block}
\vspace{-3mm}
\end{figure}

\textbf{-- LFB NL:}
Our default feature bank operator $\ltmnl(S_t, \tilde{L}_t)$ is an attention operator. 
Intuitively, we use $S_t$ to attend to features in $\tilde{L}_t$, and add the attended information back to $S_t$ via a shortcut connection.
We use a simple implementation in which $\ltmnl(S_t,\tilde{L}_t)$ is a stack of up to three non-local (NL) blocks~\cite{nonlocal}. 
We replace the self-attention of the standard non-local block~\cite{nonlocal} with attention between the local features $S_t$ and the long-term feature window $\tilde L_t$, illustrated in \figref{nl_block}.
In addition, our design uses layer normalization (LN)~\cite{ba2016layer} and dropout~\cite{srivastava2014dropout} to improve regularization.
We found these modifications to be important since our target tasks contain relatively few training videos and exhibit overfitting.
The stack of modified non-local blocks, denoted as $\nlmod$, is iterated as:\vspace{-1mm}
\begin{align}
  S^{(1)}_t &= \nlmod_{\theta_1}(S_t, \tilde{L}_t), \nonumber\\
  S^{(2)}_t &= \nlmod_{\theta_2}(S^{(1)}_t, \tilde{L}_t), \nonumber\\[-0.6em]
  &\quad\vdots\nonumber\\[-2em]\nonumber
\end{align}
where $\theta_{\{1,2, \dots\}}$ are learnable parameters. 
Similar to Wang~\etal~\cite{strg}, we use a linear layer to reduce the $\ltmnl$ input dimensionality to 512 and apply dropout~\cite{srivastava2014dropout} with rate 0.2.
Thus the input of the final linear classifier is 2048 ($S_t$) + 512 ($\ltmnl$ output) = 2560 dimensional.

%##################################################################################################
%##################################################################################################

\textbf{-- LFB Avg/Max:} This version uses a simpler variant in which $\ltmpool(S_t, \tilde{L}_t) = \mathrm{pool}(\tilde{L}_t)$, where $\mathrm{pool}$ can be either average or max pooling. 
This implementation results in a classifier input that is 2048 ($S_t$) + 2048 ($\ltmpool$ output) = 4096 dimensional.

\vspace{-0.5mm}
\paragraph{Training.}
Joint, end-to-end training of the entire model (\figref{model:ltm}) is not feasible due to the computational and memory complexity of back-propagating through the long-term feature bank. Instead, we treat the 3D CNN and detector that are used to compute $L$ as \emph{fixed} components that are trained offline, but still on the target dataset, and not updated subsequently. We have experimented with alternating optimization methods for updating these models, similar to target propagation~\cite{lecun1986learning}, but found that they did not improve results. Dataset-specific training details are given later.

\vspace{-0.5mm}
\paragraph{A Baseline Short-Term Operator.}
To validate the benefit of incorporating long-term information, 
we also study a `degraded' version of our model that does not use a long-term feature bank.
Instead, it uses a \emph{short-term operator} that is identical to $\ltmnl$, but only references the information from within a clip: $\stm(S_t) := \ltmnl(S_t,S_t)$.
$\stm$ is conceptually similar to~\cite{strg} and allows for back-propagation.
We observed substantial overfitting with $\stm$ and thus applied additional regularization techniques.
See the Appendix for details.

%%%%%%%%%%%%%%%%%%%%%%%%%%%%%%%%%%%%%%%%%%%%%%%%%%%%%%%%%%%%%%%%%%%%%%%%%%%%%%%%%%%%%%%%%%%%%%%%%%%
\section{Experiments on AVA}
\label{sec:exp_ava}

We use the AVA dataset~\cite{ava} for extensive ablation studies. AVA consists of 235 training videos and 64 validation videos; each video is a 15 minute segment taken from a movie. Frames are labeled sparsely at 1 FPS\@. The labels are: one bounding box around each person in the frame together with a multi-label annotation specifying which actions the person in the box is engaged in within $\pm$0.5 seconds of the labeled frame. The action label space is defined as 80 `atomic' actions defined by the dataset authors.

The task in AVA is spatio-temporal action localization: each person appearing in a test video must be detected in each frame and the multi-label actions of the detected person must be predicted correctly. The quality of an algorithm is judged by a mean average precision (mAP) metric that requires at least 50\% intersection over union (IoU) overlap for a detection to be matched to the ground-truth while simultaneously predicting the correct actions.

%##################################################################################################
\begin{table*}[t]\vspace{-4mm}
\subfloat[\textbf{Temporal support} (mAP in \%)\label{tab:ablation:temporal}]{%
\tablestyle{2.5pt}{1.1}\begin{tabular}{@{}lx{20}x{20}x{20}x{20}x{20}x{20}x{20}x{20}@{}}
  Support (sec.)& 2 & 3 & 5 & 10 & 15 & 30 & 60\\
\shline
 {3D CNN} & 22.1 & 22.2 & 22.3 & 20.0 & 19.7 & 17.5 & 15.7\\
 {STO} & 23.2 & 23.6 & 23.3 & 21.5 & 20.9 & 18.5 & 16.9\\
 {\bf LFB (causal)} & - & 24.0 & 24.3 & 24.6 & 24.8 & 24.6 & 24.2\\
 {\bf LFB (batch)} & - & 24.2 & 24.7 & 25.2 & \underline{25.3} & \underline{25.3} & \bf25.5\\
\end{tabular}}\hfill
\subfloat[\textbf{Feature decoupling}\label{tab:ablation:decoupling}]{%
\tablestyle{4.8pt}{1.1}\begin{tabular}{@{}lx{20}@{}}
 & mAP\\
\shline
 {K400 feat.\ \ \ \ \ \ \ } & 19.7\\
 {AVA feat.} & \underline{24.3}\\
 {\bf LFB} & \bf 25.5\\
 \multicolumn{2}{c}{~}\\
\end{tabular}}\hfill
\subfloat[\textbf{LFB operator}\label{tab:ablation:function}]{%
\tablestyle{4.8pt}{1.1}\begin{tabular}{@{}lx{20}@{}}
 & mAP\\
\shline
 {Avg pool\ \ \ } & 23.1\\
 {Max pool} & \underline{23.2}\\
 {\bf NL} & \bf 25.5\\
 \multicolumn{2}{c}{~}\\
\end{tabular}}\hfill
\subfloat[\textbf{LFB spatial design}\label{tab:ablation:input}]{%
\tablestyle{4.8pt}{1.1}\begin{tabular}{@{}lx{20}@{}}
 & mAP \\
\shline
 {Global pool\ \ \ \ \ \ } & 24.9\\
 {$2\x2$ Grid} & \underline{25.1}\\
 {$4\x4$ Grid} & \underline{25.1}\\
 {\textbf{Detection}} & \bf 25.5\\
\end{tabular}}\\
\subfloat[\textbf{LFB NL design}\label{tab:ablation:nlblock}]{%
\tablestyle{4.8pt}{1.1}\begin{tabular}{@{}lx{19}@{}}
  & mAP\\
\shline
  {1L} & 25.1\\
  {\bf 2L (default)} & \underline{25.5}\\
  {2L w/o scale} & 25.2\\
  {2L w/o LN} & 23.9\\
  {2L w/o dropout} & 25.4\\
  {2L (dot product)} & \underline{25.5}\\
  {2L (concat)} & 25.3\\
  {3L} & \bf25.8\\
 \multicolumn{2}{c}{~}\\
\end{tabular}}\hfill
\subfloat[\textbf{Model complexity}\label{tab:ablation:complexity}]{%
\tablestyle{4pt}{1.1}\begin{tabular}{@{}lx{19}x{19}x{19}@{}}
 & params & FLOPs & mAP\\
\shline
 {3D CNN} & $1\x$ & $1\x$ & 22.1\\
 {3D CNN $\times$2} & $2\x$ & $2\x$ & 22.9\\
 {STO} & $1.00\x$ & $1.12\x$ & 23.2\\
 {STO $\times$2} & $2.00\x$ & $2.24\x$ & 24.1\\
 {\textbf{LFB (2L)}} & $2.00\x$ & $2.12\x$ & \underline{25.5}\\
 {\textbf{LFB (3L)}} & $2.00\x$ & $2.15\x$ & \bf 25.8\\
 \multicolumn{2}{c}{~}\\
 \multicolumn{2}{c}{~}\\
 \multicolumn{2}{c}{~}\\
\end{tabular}}\hfill
\subfloat[\textbf{Backbone \& testing}\label{tab:ablation:r101}]{%
\tablestyle{4.8pt}{1.1}\begin{tabular}{@{}lx{19}@{}}
  & mAP\\
\shline
 \bf R50-I3D-NL\\
  \quad center-crop (default) & 25.8\\
 \bf R101-I3D-NL\\
  \quad center-crop (default) & 26.8\\
  \quad 3-crop & 27.1\\
  \quad 3-crop+flips & \underline{27.4}\\
  \quad 3-crop+flips+3-scale & \bf 27.7\\
 \multicolumn{2}{c}{~}\\
 \multicolumn{2}{c}{~}\\
\end{tabular}}\hfill
\subfloat[\textbf{Comparison to prior work}\label{tab:ablation:priorwork}]{%
\tablestyle{4.8pt}{1.1}\begin{tabular}{@{}l|x{13}|x{19}x{19}@{}}
 model & flow & val & test\\
\shline
 {AVA} \cite{ava} & \checkmark & 15.6 & -\\
 {ACRN} \cite{sun2018actor} & \checkmark & 17.4 & -\\
 {RTPR} \cite{li2018recurrent} & \checkmark & 22.3 & -\\
 {\demph{9-model ens.}\cite{jianghuman}} & \checkmark & \demph{25.6} & \demph{21.1}\\
\hline
 {R50-I3D-NL} \cite{jianghuman} & & 19.3 & -\\
 {RTPR} \cite{li2018recurrent} & & 20.5 & -\\
 {Girdhar \etal} \cite{girdhar2018better} & & 21.9 & 21.0\\
 {\textbf{LFB}} (R50) & & \underline{25.8} & \underline{24.8}\\
 {\textbf{LFB} (best R101)} &  & {\bf 27.7} & {\bf 27.2}\\
\end{tabular}}

\vspace{1mm}
\caption{\textbf{AVA ablations and test results}. \textbf{STO:} 3D CNN with a non-local (NL) short-term operator; \textbf{LFB:} 3D CNN with a \textbf{long-term feature bank}; the LFB operator is a two-layer (2L) NL block by default. We perform ablations on the AVA spatio-temporal action localization. The results validate that longer-term information is beneficial, that the improvement is larger than what would be observed by ensembling, and demonstrate various design choices. Finally, we show state-of-the-art results on the AVA test set.}
\label{tab:ablations}\vspace{-3mm}
\end{table*}
%##################################################################################################

\subsection{Implementation Details}
\label{sec:exp_ava_impl}
Next, we describe the object detector, input sampling, and training and inference details used for AVA.

\paragraph{Person Detector.}
We use Faster R-CNN~\cite{Ren2015a} with a ResNeXt-101-FPN~\cite{xie2017aggregated,lin2017feature} backbone for person detection.
The model is pre-trained on ImageNet~\cite{Russakovsky2015} and COCO keypoints~\cite{lin2014microsoft}, and then fine-tuned on AVA bounding boxes;
see the Appendix for training details.
The final model obtains 93.9 AP@50 on the AVA validation set.

\paragraph{Temporal Sampling.} Both short- and long-term features are extracted by 3D CNNs that use 32 input frames sampled with a temporal stride of 2 spanning 63 frames ($\sim$2 seconds in 30 FPS video). Long-term features are computed at one clip per second over the whole video, with a 3D CNN model (\figref{model:baseline}) fine-tuned on AVA\@.

\paragraph{Training.}
We train our models using synchronous SGD with a minibatch size of 16 clips on 8 GPUs (\ie, 2 clips per GPU), with
batch normalization~\cite{ioffe2015batch} layers frozen.
We train all models for 140k iterations, with a learning rate of 0.04, which is decreased by a factor of 10 
at iteration 100k and 120k. 
We use a weight decay of $10^{-6}$ and momentum of 0.9. 
For data augmentation, we perform random flipping, 
random scaling such that the short side $\in [256, 320]$ pixels, 
and random cropping of size 224$\x$224.
We use both ground-truth boxes and predicted boxes with scores at least 0.9 for training.
This accounts for the discrepancy between ground-truth-box distribution and predicted-box distribution, and we found it beneficial.
We assign labels of a ground-truth box to a predicted box if they overlap with IoU at least 0.9.
A predicted box might have no labels assigned. 
Since the number of long-term features $N$ differs from clip to clip, we pad zero-vectors for clips with 
fewer long-term features to simplify minibatch training.

\paragraph{Inference.} 
At test time we use detections with scores $\ge$ 0.85. 
All models rescale the short side to 256 pixels and use a single center crop of 256$\x$256.
For both training and inference, if a box crosses the cropping boundary, we pool the region within the cropped clip.
In the rare case where a box falls out of the cropped region, RoIAlign~\cite{He2017} pools the feature at the boundary.

\subsection{Ablation Experiments}

%##################################################################################################
\begin{figure*}[t!]
\centering
\includegraphics[width=1.0\linewidth, page=1]{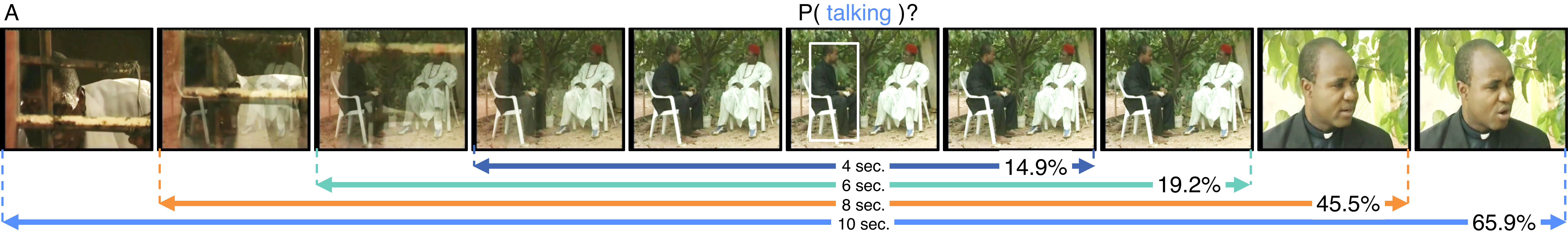}
\includegraphics[width=1.0\linewidth, page=2]{figs/window_viz2_small.pdf}\vspace{-3mm}
\includegraphics[width=1.0\linewidth, page=3]{figs/window_viz2_small.pdf}\vspace{-0.5mm}
\includegraphics[width=1.0\linewidth, page=4]{figs/window_viz2_small.pdf}
\caption{\textbf{Example Predictions.}
We compare predictions made by models using LFB of different window sizes. 
Through LFB, the model is able to exploit information distant in time, \eg, the zoomed-in frames in example A and C, to improve predictions.
We encourage readers to zoom in for details.  
({\color{fbblue1}Blue}: correct labels. {\color{fbred3}Red}: incorrect labels. Best viewed on screen.)
}
\vspace{-3mm}
\label{fig:viz}
\end{figure*}
%##################################################################################################

\paragraph{Temporal Support.}
We first analyze the impact of increasing temporal support on models both with and without LFB\@.
For models without LFB, we evaluate a vanilla 3D CNN (\figref{model:baseline}) and a 3D CNN extended with STO (denoted `STO' in tables).
To increase their temporal support, we increase the temporal stride
but fix the number of input frames
so that the model covers a longer temporal range while still being feasible to train.
To increase the temporal support of LFB models, we increase the `window size' $2w + 1$ of $\tilde{L}_t$.

\tabref{ablation:temporal} compares model performance.
Increasing temporal support through striding in fact hurts the performance of both `3D CNN' and `STO'.
Temporal convolution might not be suitable for long-term patterns, 
since long-term patterns are more diverse, and include challenging scene cuts.
On the other hand,
increasing temporal support by adding an LFB steadily improves performance, 
leading to a large gain over the original `3D CNN' (22.1 $\rightarrow$ {\bf25.5}).
The online (causal) setting shows a similar trend.

Overall we observe strong improvements given by long-range context even though AVA actions are designed to be `atomic' and localized within $\pm$0.5 seconds.
For the rest of the ablation study, we focus on the batch setting and use a window size of 60 seconds due to the strong performance. 

\paragraph{Feature Decoupling.}
In \tabref{ablation:decoupling}, we compare our decoupled feature approach with prior long-term modeling strategies where a single, pre-computed feature type is used (\eg~\cite{yue2015beyond,li2017temporal,miech2017learnable,tang2018non}). To do this, we use the same 3D CNN for both short-term and long-term feature bank computation and keep it fixed during training; only the parameters of the FBO and classifier are updated. We consider two choices: a Kinetics-400~\cite{i3d} pre-trained 3D CNN (`K400 feat.') and a 3D CNN that is fine-tuned on AVA (`AVA feat.'). Our decoupled approach, which updates the short-term 3D CNN based on the long-term context, works significantly better. 

\paragraph{FBO Function Design.}
We next compare different $\ltm$ function designs in \tabref{ablation:function}.
We see that a non-local function significantly outperforms pooling on AVA\@.
This is not surprising, since videos in AVA (and videos in general) are
multi-actor, multi-action, and can exhibit complex interactions possibly across a long range of time.
We expect a more complex function class is required for reasoning through the complex scenes.
Nonetheless, pooling still offers a clear improvement.
This again confirms the importance of long-term context in video understanding. 

\paragraph{FBO Input Design.}
What spatial granularity is required for complex video understanding? 
In \tabref{ablation:input}, we compare constructing long-term features 
using detected objects (`Detection'), regular grids (`Grid'), 
and non-spatial features (`Global pool'). 
In the `Grid' experiments, we divide the feature map into a $k \x k$ grid, for $k = 2,4$, and average the features within each bin to obtain a localized representation without an object detector (similar to ACRN~\cite{sun2018actor}). 

\tabref{ablation:input} shows that the actor-level features (`Detection')
works better than coarser regular-grid or non-spatial features.
We believe this suggests a promising future research direction that moves from global pooling 
to a more detailed modeling of objects/actors in video. 

\vspace{-1mm}
\paragraph{Non-Local Block Design.}
Next, we ablate our NL block design. 
In \tabref{ablation:nlblock}, we see that adding a second layer of NL blocks leads to improved performance.
In addition, scaling~\cite{vaswani2017attention}, layer normalization~\cite{ba2016layer}, 
and dropout~\cite{srivastava2014dropout} all contribute to good performance. 
Layer normalization~\cite{ba2016layer}, which is part of our modified NL design, is particularly important. As found in~\cite{nonlocal}, the default embedded Gaussian variant performs similarly to dot product and concatenation. (Later we found that adding a third layer of NL blocks further improves accuracy, but otherwise use two by default.)

\vspace{-1mm}
\paragraph{Model Complexity.}
Our approach uses two instances of the backbone model: one to compute the long-term features and another to compute the short-term features.
It thus uses around 2$\x$ more parameters and computation than our baselines.
What if we simply use 2$\x$ more computation on baselines through an ensemble?
We find that both `3D CNN' or `STO' are not able to obtain a similar gain when ensembled;
the LFB model works significantly better (\tabref{ablation:complexity}).

\vspace{-1mm}
\paragraph{Example Predictions.}
We qualitatively present a few examples illustrating the impact of LFB in \figref{viz}. 
Specifically, we compare predictions made by models with LFB of different window sizes, from 4 seconds to 10 seconds.
We see that when observing only short-term information, the model is confused and not able to make accurate predictions in these cases. 
When observing more context, \eg, zoomed-in frames (example A, C) or frames that give clearer cue (example B, D),
an LFB model is able to leverage the information and improve predictions.

\vspace{-1mm}
\paragraph{Backbone and Testing.}
So far for simplicity, we have used a relatively small backbone of R50-I3D-NL and performed center-crop testing.
In \tabref{ablation:r101}, we show that with an R101-I3D-NL backbone, LFB (3L) achieves 26.8 mAP, 
and with standard testing techniques, 27.7 mAP.
We use short side $\in \cbr{224, 256, 320}$ pixels for 3-scale testing.

\vspace{-1mm}
\paragraph{Comparison to Prior Work.}
Finally we compare with other state-of-the-art methods (\tabref{ablation:priorwork}).
For fair comparison, we follow Girdhar \etal \cite{girdhar2018better} and train on both the training and validation set
for test set evaluation.\footnote{Test set performance evaluated by ActivityNet server.}
For this model we use a 1.5$\times$ longer schedule due to the larger data size.

Our model, using only RGB frames, significantly outperforms all prior work
including strong competition winners that use optical flow and large ensembles.
Our single model outperforms the best previous single-model entry by Girdhar \etal~\cite{girdhar2018better} with a margin of 5.8 and 6.2 points mAP on validation and test set respectively. 

%%%%%%%%%%%%%%%%%%%%%%%%%%%%%%%%%%%%%%%%%%%%%%%%%%%%%%%%%%%%%%%%%%%%%%%%%%%%%%%%%%%%%%%%%%%%%%%%%%%
\section{Experiments on EPIC-Kitchens}
\label{sec:exp_epic}
\vspace{-0.5mm}

The long-term feature bank is a generalizable and flexible concept. We illustrate this point on two tasks in the EPIC-Kitchens dataset~\cite{epic} for which we store different types of information in the feature bank.

The EPIC-Kitchens dataset consists of videos of daily activities (mostly cooking) recorded in
participants' native kitchen environments. 
Segments of each video are annotated with one verb (\eg, `squeeze') and one noun (\eg, `lemon').
The task is to predict the \emph{verb}, \emph{noun}, and the combination (termed \emph{action} %
\cite{epic}) in each segment. 
Performance is measured by top-1 and top-5 accuracy. 

The dataset consists of 39,594 segments in 432 videos.
Test set annotations are not released;
for validation we split the original training set into a new training/validation split following Baradel \etal~\cite{baradel2018object}.  
We train independent models to recognize \emph{verbs} and \emph{nouns},
and combine their predictions for \emph{actions}.
For actions, we additionally use a prior based on the training class frequencies of verbs and nouns; see the Appendix for details.

\vspace{-0.5mm}
\subsection{Implementation Details}
\vspace{-0.5mm}

\paragraph{Long-Term Feature Banks.}
Recognizing which object a person is interacting with (the \emph{noun} task) from a short segment is challenging,
because the object is often occluded, blurred, or can even be out of the scene.
Our LFB is well-suited for addressing these issues, as the long-term supporting information can help resolve ambiguities.
For example, if we know that the person took a lemon from the refrigerator 30 seconds ago,
then cutting a lemon should be more likely.
Based on this motivation, we construct an LFB that contains \emph{object-centric features}.
Specifically, we use Faster R-CNN to detect objects and extract object features using RoIAlign from the detector's feature maps
(see the Appendix for details of our detector).

On the other hand, for recognizing \emph{verbs} %
we use a \emph{video model} to construct an LFB that captures motion patterns.
Specifically, we use our baseline 3D CNN fine-tuned on EPIC-Kitchens verbs to extract clip-level features every 1 second of video.
Our default setting uses a window size of 40 seconds for the \emph{verb} model, 
and 12 seconds for the \emph{noun} model, chosen on the validation set.

\vspace{-0.5mm}
\paragraph{Adaptation to Segment-Level Tasks.}
To adapt our model for segment-level predictions, we replace the RoI pooling by
global average pooling, resulting in $S \in \RR^{1\times2048}$.
For the $\stm$ baseline, we use a slightly modified formulation: $\stm(S'_t) := \ltm_{\textrm{NL}}(S_t, S'_t)$,
where $S'_t$ contains 16 spatially pooled features, each at a temporal location, and $S_t$ is $S'_t$ pooled over the time axis.
$\stm$ learns to interact with information at different time steps within the short clip.
Training and inference procedures are analogous to our experiments on AVA; see the Appendix for details.

%##################################################################################################
\begin{table}[tb]
\resizebox{\columnwidth}{!}{\tablestyle{4.5pt}{1.1}\begin{tabular}{@{}lx{22}x{22}x{22}x{22}x{22}x{22}@{}}
 & \multicolumn{2}{c}{Verbs} & \multicolumn{2}{c}{Nouns} & \multicolumn{2}{c}{Actions}\\
 \cline{2-3}
 \cline{4-5}
 \cline{6-7}
  & top-1 & top-5 & top-1 & top-5 & top-1 & top-5\\
\shline
 \textbf{Validation}\\
 {\quad Baradel~\cite{baradel2018object}} & 40.9 & - & - & - & - & -\\
 {\quad 3D CNN} & 49.8 & 80.6 & 26.1 & 51.3 & 19.0 & 37.8\\
 {\quad 3D CNN ens.} & 50.7 & \underline{81.2} & 27.8 & 52.8 & 20.0 & 39.0\\
 {\quad STO} & 51.0 & 80.8 & 26.6 & 51.5 & 19.5 & 38.3\\
 {\quad STO ens.} & 51.9 & \underline{81.2} & 27.8 & 52.5 & 20.5 & 39.4\\
 {\quad \textbf{LFB NL}} & 51.7  & \underline{81.2} & \underline{29.2} & 55.3 & \underline{21.4} & 40.2\\
 {\quad \textbf{LFB Avg}} & \bf{53.0} & \bf{82.3} & 29.1 & \underline{55.4} & 21.2 & \underline{40.8}\\
 {\quad \textbf{LFB Max}} & \underline{52.6}  & \underline{81.2} & \bf 31.8 & \bf 56.8 & \bf 22.8 & \bf 41.1\\
 \quad\quad\ \ $\Delta$ & \emph{+3.2} & \emph{+1.7} & \emph{+5.7} & \emph{+5.5} & \emph{+3.8} & \emph{+3.3}\\
 \hline
 \textbf{Test s1 (seen)}\\
 {\quad TSN RGB}~\cite{epic} & 45.7 & 85.6 & 36.8 & 64.2 & 19.9 & 41.9\\
 {\quad TSN Flow}~\cite{epic} & 42.8 & 79.5 & 17.4 & 39.4 & 9.0 & 21.9\\
 {\quad TSN Fusion}~\cite{epic} & 48.2 & 84.1 & 36.7 & 62.3 & 20.5 & 39.8\\
 {\quad \textbf{LFB Max}} & \bf 60.0 & \bf 88.4 & \bf 45.0 & \bf 71.8 & \bf 32.7 & \bf 55.3\\
 \hline
 \textbf{Test s2 (unseen)}\\
 {\quad TSN RGB}~\cite{epic} & 34.9 & 74.6 & 21.8 & 45.3 & 10.1 & 25.3\\
 {\quad TSN Flow}~\cite{epic} & 40.1 & 73.4 & 14.5 & 33.8 & 6.7 & 18.6\\
 {\quad TSN Fusion}~\cite{epic} & 39.4 & 74.3 & 22.7 & 45.7 & 10.9 & 25.3\\
 {\quad \textbf{LFB Max}} & \bf 50.9 & \bf 77.6 & \bf 31.5 & \bf 57.8 & \bf 21.2 & \bf 39.4\\
\end{tabular}}
\vspace{1mm}
  \caption{\textbf{EPIC-Kitchens validation and test server results.}
  Augmenting 3D CNN with LFB leads to significant improvement.
  }\label{tab:epic}
\vspace{-3mm}
\end{table}
%##################################################################################################

\subsection{Quantitative Evaluation}
We now quantitatively evaluate our LFB models.
\tabref{epic} shows that augmenting a 3D CNN with LFB significantly boosts the performance for all three tasks.
Using object features for the \emph{noun} model is particularly effective, 
leading to 5.7\% (26.1 $\rightarrow$ {\bf 31.8}) absolute improvement over our strong baseline model.
On \emph{verb} recognition, LFB with 3D CNN features results in 3.2\% (49.8 $\rightarrow$ {\bf 53.0}) improvement and outperforms previous state-of-the-art by Baradel \etal~\cite{baradel2018object}
by 12.1\% (40.9 $\rightarrow$ {\bf 53.0}).

We also observe that $\ltmmax$ and $\ltmavg$ outperform $\ltmnl$ on EPIC-Kitchens.
We conjecture that this is due to the simpler setting:
each video has only one person, doing one thing at a time, without the complicated person-person interactions of AVA\@.
Thus a simpler function suffices.

On the test set, 
our method outperforms prior work by a large margin on both the `seen kitchens (s1)' and 
`unseen kitchens (s2)' settings. 
Our LFB model outperforms the Two-Stream~\cite{simonyan2014two} TSN~\cite{tsn} baseline by 50\% relatively for s1, and almost doubles the performance for s2, in terms of top-1 \emph{action} accuracy. 

%%%%%%%%%%%%%%%%%%%%%%%%%%%%%%%%%%%%%%%%%%%%%%%%%%%%%%%%%%%%%%%%%%%%%%%%%%%%%%%%%%%%%%%%%%%%%%%%%%%
\vspace{-0.5mm}
\section{Experiments on Charades}
\vspace{-0.5mm}
\label{sec:exp_charades}

Finally we evaluate our approach on the Charades dataset~\cite{charades}. 
The Charades dataset contains 9,848 videos with an average length of 30 seconds. 
In each video, a person can perform one or more actions. 
The task is to recognize all the actions in the video without localization.

\vspace{-0.5mm}
\subsection{Implementation Details}
\vspace{-0.5mm}
We use the RGB frames at 24 FPS provided by the dataset authors. 
We sample training and testing clips (32 frames) with a temporal stride of 4 following STRG~\cite{strg},
resulting in input clips spanning 125 frames ($\sim$5.2 seconds).
The LFB is sampled at 2 clips per second.
We found a post-activation version of $\nlmod$ to work better on Charades, so we adopt it in the following experiments.
Details and full results of both variants are in the Appendix.
Other details are identical to the verb model for EPIC-Kitchens.

%##################################################################################################
\begin{table}[t]
\resizebox{\columnwidth}{!}{\tablestyle{8pt}{1.1}\begin{tabular}{@{}lrr@{}}
 iterations / lr / wd & 50k / 0.0025 / 1e-4 \cite{strg}& 24k / 0.02 / 1.25e-5\\
\shline
{3D CNN} & 33.8 & \bf 38.3\\
{STO} & 37.8 & \bf 39.6\\
\end{tabular}}
\vspace{0.5mm}
\caption{%
  \textbf{Training schedule on Charades.} 
  Our 2$\x$ shorter schedule works significantly better than the schedule used in STRG~\cite{strg}. 
}\label{tab:charades_schedule}
\vspace{-2mm}
\end{table}
%##################################################################################################

%##################################################################################################
\begin{table}[t]
\resizebox{0.99\columnwidth}{!}{%
\tablestyle{3.3pt}{1.1}\begin{tabular}{@{}lllx{22}x{22}@{}}
  & backbone & modality & {train} / {val} & {trainval} / {test}\\
\shline
2-Strm. \cite{simonyan2014two} (from \cite{sigurdsson2017asynchronous})  & VGG16 & RGB+Flow & 18.6 & -\\
Asyn-TF \cite{sigurdsson2017asynchronous} & VGG16 & RGB+Flow & 22.4 & -\\
CoViAR \cite{coviar} & R50 & Compressed & 21.9 & -\\
MultiScale TRN~\cite{zhou2017temporalrelation} & Inception & RGB & 25.2 & -\\
I3D \cite{i3d} & Inception-I3D & RGB & 32.9 & 34.4\\
I3D \cite{i3d} (from \cite{nonlocal}) & R101-I3D & RGB & 35.5 & 37.2\\
I3D-NL \cite{nonlocal} (from \cite{strg}) & R50-I3D-NL & RGB & 33.5 & -\\
I3D-NL \cite{nonlocal} & R101-I3D-NL & RGB & 37.5 & 39.5\\
STRG \cite{strg} & R50-I3D-NL & RGB & 37.5 & -\\
STRG \cite{strg} & R101-I3D-NL & RGB & 39.7 & -\\
\hline
{3D CNN} & R50-I3D-NL & RGB &38.3 & -\\
{3D CNN ens.} & R50-I3D-NL & RGB & 39.5 & -\\
{STO} & R50-I3D-NL & RGB & 39.6 & -\\
{STO ens.} & R50-I3D-NL & RGB & 40.0 & -\\
\textbf{LFB Avg} & R50-I3D-NL & RGB & 38.4 & -\\
\textbf{LFB Max} & R50-I3D-NL & RGB & 38.6 & -\\
\textbf{LFB NL} & R50-I3D-NL & RGB & \bf40.3 & -\\
\hline
{3D CNN} & R101-I3D-NL & RGB & 40.3 & 40.8\\
{3D CNN ens.} & R101-I3D-NL & RGB & 41.7 & -\\
{STO} & R101-I3D-NL & RGB & 41.0 & -\\
{STO ens.} & R101-I3D-NL & RGB & 42.3 &-\\
\textbf{LFB Avg} & R101-I3D-NL & RGB & 40.8 & -\\
\textbf{LFB Max} & R101-I3D-NL & RGB & 40.9 & -\\
\textbf{LFB NL} & R101-I3D-NL & RGB & \bf42.5 & \bf43.4\\
\end{tabular}
}
\vspace{0.5mm}
\caption{\textbf{Action recognition accuracy on Charades.} (mAP in \%)}\label{tab:charades}
\vspace{-3.5mm}
\end{table}
%##################################################################################################

\vspace{-0.25mm}
\paragraph{Training and Inference.}
\vspace{-0.25mm}
We train 3D CNN models for 24k iterations with a learning rate of 0.02
and weight decay of 1.25e-5.
Note that these hyperparameters are different from STRG~\cite{strg}, 
which uses longer schedule (50k iterations), smaller learning rate (0.0025),
and larger weight decay (1e-4).\footnote{The original STRG~\cite{strg} uses a batch size of 8. For clear comparison, 
we use the same batch size as ours (16), but adjust the learning rate and schedule according to `Linear Scaling Rule'~\cite{goyal2017imagenet1hr}.
We verified that the accuracy matches that of the original 4-GPU training.} 
\tabref{charades_schedule} compares the two settings, and we see that surprisingly our 2$\x$ \emph{shorter} schedule works significantly better.
With the new schedule, a simple NL model without proposals (STO) works as well as the full STRG method (37.5\% mAP)~\cite{strg}.
We observe that the benefit of using the short-term operator becomes smaller when using a stronger baseline.
In all following experiments we use our 24k schedule as default,
and use a 2-stage training approach similar to STRG~\cite{strg} for training LFB models; see the Appendix for details.
At test time, we sample 10 clips per video, and combine the predictions using max pooling
following prior work~\cite{nonlocal, strg}.
We use (left, center, right) 3-crop testing following Wang~\etal~\cite{nonlocal}.

\subsection{Quantitative Evaluation}
For Charades, we experiment with both ResNet-50-I3D-NL and ResNet-101-I3D-NL~\cite{He2016,i3d,nonlocal} backbones for a consistent comparison to prior work.
\tabref{charades} shows that
LFB models again consistently outperform all models without LFB,
including prior state-of-the-art
on both validation and test sets.
The improvement on Charades is not as large as other datasets, in part due to the coarser prediction task (video-level).

%%%%%%%%%%%%%%%%%%%%%%%%%%%%%%%%%%%%%%%%%%%%%%%%%%%%%%%%%%%%%%%%%%%%%%%%%%%%%%%%%%%%%%%%%%%%%%%%%%%
\vspace{-1mm}
\section{Discussion}\label{sec:discussion}
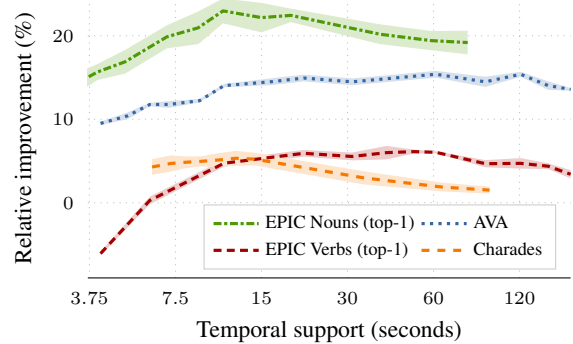
\begin{figure}[t]
\tablestyle{4.8pt}{1.05}\begin{tabular}{@{}c@{}}
  \small
  \begin{tikzpicture}
  \begin{axis}[
    xmode=log,
    xtick={3.75, 7.5,15,30,60,120},
    log ticks with fixed point,
    legend pos=south east,
    legend style={font=\scriptsize},
    legend columns=2,
    width=8cm,
    height=5.3cm,
    xmin=3.7,
    ymin=-9,
    axis lines=left,
    y axis line style={draw opacity=0},
    xlabel={Temporal support (seconds)},
    ylabel={Relative improvement (\%)},
    every axis plot/.append style={very thick},
      cycle multiindex* list={%
        mycolorlist
            \nextlist
        mystylelist
      },
  ]
  \pgfplotsset{every tick label/.append style={font=\scriptsize}}
  \addplot+[mark options={solid}]
      table {plotdata/epic_noun_window_max_top1_gain.data};
      \addlegendentry{EPIC Nouns (top-1)}
  \addplot+[mark options={solid}]
      table {plotdata/ava_window_nl_gain.data};
      \addlegendentry{AVA}
  \addplot+[mark options={solid}]
      table {plotdata/epic_verb_window_max_top1_gain.data};
      \addlegendentry{EPIC Verbs (top-1)}
  \addplot+[mark options={solid}]
      table {plotdata/charades_window_nl_gain.data};
      \addlegendentry{Charades}
  \addplot [name path=upper,draw=none] table[x=x,y expr=\thisrow{y}+\thisrow{stderr}] {plotdata/epic_noun_window_max_top1_gain.data};
  \addplot [name path=lower,draw=none] table[x=x,y expr=\thisrow{y}-\thisrow{stderr}] {plotdata/epic_noun_window_max_top1_gain.data};
  \addplot [fill=chameleon3!20] fill between[of=upper and lower];
  \addplot [name path=upper,draw=none] table[x=x,y expr=\thisrow{y}+\thisrow{stderr}] {plotdata/ava_window_nl_gain.data};
  \addplot [name path=lower,draw=none] table[x=x,y expr=\thisrow{y}-\thisrow{stderr}] {plotdata/ava_window_nl_gain.data};
  \addplot [fill=skyblue2!20] fill between[of=upper and lower];
  \addplot [name path=upper,draw=none] table[x=x,y expr=\thisrow{y}+\thisrow{stderr}] {plotdata/epic_verb_window_max_top1_gain.data};
  \addplot [name path=lower,draw=none] table[x=x,y expr=\thisrow{y}-\thisrow{stderr}] {plotdata/epic_verb_window_max_top1_gain.data};
  \addplot [fill=scarletred3!20] fill between[of=upper and lower];
  \addplot [name path=upper,draw=none] table[x=x,y expr=\thisrow{y}+\thisrow{stderr}] {plotdata/charades_window_nl_gain.data};
  \addplot [name path=lower,draw=none] table[x=x,y expr=\thisrow{y}-\thisrow{stderr}] {plotdata/charades_window_nl_gain.data};
  \addplot [fill=orange2!20] fill between[of=upper and lower];
  \end{axis}
  \end{tikzpicture}
  \end{tabular}\vspace{-1mm}
\caption{\textbf{Relative improvement} of LFB models with different window sizes over vanilla 3D CNN.}\label{fig:summary:improv}
\vspace{-3.5mm}
\end{figure}

\vspace{-1mm}
\figref{summary:improv} shows the relative gain of using LFB
of different window sizes.\footnote{For each dataset, we use its best-performing $\ltm$. Standard error is calculated based on 5 runs.
The temporal support here considers the support of each clip used for computing $L$,
so Charades's support starts at a higher value due to its larger 3D CNN clip size ($\sim$5.2 seconds).}
We see that different datasets exhibit different characteristics.
The \emph{movie} dataset, AVA, benefits from very long context lasting 2+ minutes.
To recognize \emph{cooking} activities (EPIC-Kitchens), context spanning from 15 to 60 seconds is useful.
Charades videos are much shorter ($\sim$30 seconds),
but still, extending the temporal support to 10+ seconds is beneficial. 
We conjecture that more challenging datasets in the future may benefit even more.

%##################################################################################################
%##################################################################################################

In conclusion, we propose a \textbf{Long-Term Feature Bank} that provides long-term supportive information to video models.
We show that enabling video models with access to long-term information, through an LFB, leads to a large performance gain and yields state-of-the-art results on challenging datasets like AVA, EPIC-Kitchens, and Charades.

\begin{appendices}

%%%%%%%%%%%%%%%%%%%%%%%%%%%%%%%%%%%%%%%%%%%%%%%%%%%%%%%%%%%%%%%%%%%%%%%%%%%%%%%%%%%%%%%%%%%%%%%%%%%
\section{Backbone Architecture}
We use ResNet-50 I3D~\cite{He2016,ava} with non-local blocks~\cite{nonlocal} as the `backbone' of our model.
Following Wang \etal~\cite{strg}, the network only downsamples the temporal dimension by a factor of two.
\tabref{backbone} presents the exact specification.

%##################################################################################################
\newcommand{\blockb}[3]{\multirow{3}{*}{\(\left[\begin{array}{l}\text{3$\x$1$\x$1, #2}\\[-.1em] \text{1$\x$3$\x$3, #2}\\[-.1em] \text{1$\x$1$\x$1, #1}\end{array}\right]\)$\x$#3}}
\newcommand{\blockbgeneral}[3]{\multirow{3}{*}{\(\left[\begin{array}{l}\text{3(1)$\x$1$\x$1, #2}\\[-.1em] \text{1\ \ \ \ $\x$3$\x$3, #2}\\[-.1em] \text{1\ \ \ \ $\x$1$\x$1, #1}\end{array}\right]\)$\x$#3}}
\begin{table}[h]
\footnotesize
\tablestyle{11.6pt}{1.05}
\begin{tabular}{@{}llc@{}}
Stage & \multicolumn{1}{c}{Specification} & Output size \\
\shline
conv$_1$ & \multicolumn{1}{c}{5$\x$7$\x$7, 64, stride 1, 2, 2} & 32$\x$112$\x$112 \\
\hline
pool$_1$  & \multicolumn{1}{c}{1$\x$3$\x$3 max, stride 1, 2, 2} & 32$\x$56$\x$56 \\
\hline
\multirow{3}{*}{res$_2$} & \blockb{256\ \ \ \ \ }{64\ \ \ \ \ }{3} & \multirow{3}{*}{32$\x$56$\x$56} \\
  &  & \\
  &  & \\
\hline
pool$_2$  & \multicolumn{1}{c}{2$\x$1$\x$1 max, stride 2, 1, 1} & 16$\x$56$\x$56 \\
\hline
\multirow{3}{*}{res$_3$} & \blockbgeneral{512\ }{128\ }{4, NL: 1, 3} & \multirow{3}{*}{16$\x$28$\x$28} \\
  &  & \\
  &  & \\
\hline
\multirow{3}{*}{res$_4$} & \blockbgeneral{1024}{256}{6, NL: 1, 3, 5} & \multirow{3}{*}{16$\x$14$\x$14}  \\
  &  & \\
  &  & \\
\hline
\multirow{3}{*}{res$_5$} & \blockbgeneral{2048}{512}{3} & \multirow{3}{*}{16$\x$14$\x$14} \\
  &  & \\
  &  & \\
\end{tabular}
\vspace{1mm}
\caption{ResNet-50 NL-I3D~\cite{He2016,ava,nonlocal} backbone used in this paper. 
  Here we assume input size 32$\x$224$\x$224 (frames$\x$width$\x$height).
  `NL: $i_0, i_1, \dots$' in stage `res$_j$' denotes additional non-local blocks~\cite{nonlocal} after block $i_0$, $i_1$, $\dots$ of res$_j$.
  3(1)$\x$1$\x$1 denotes that we either use a 3$\x$1$\x$1 or a 1$\x$1$\x$1 convolution.
  Specifically, we use 3$\x$1$\x$1 for block 0, 2 of res$_3$, block 0, 2, 4 of res$_4$, and block 1 of res$_5$, and use 1$\x$1$\x$1 for the rest.
}
\label{tab:backbone}
\end{table}
%##################################################################################################

%%%%%%%%%%%%%%%%%%%%%%%%%%%%%%%%%%%%%%%%%%%%%%%%%%%%%%%%%%%%%%%%%%%%%%%%%%%%%%%%%%%%%%%%%%%%%%%%%%%
\section{AVA Person Detector}
We use Faster R-CNN~\cite{Ren2015a} with a ResNeXt-101-FPN~\cite{xie2017aggregated,lin2017feature} backbone for person detection.
The model is pre-trained on ImageNet for classification, and on COCO for keypoint detection. 
The model obtains 56.9 box AP and 67.0 keypoint AP on COCO keypoints.
Model parameters are available in Detectron Model Zoo~\cite{girshick2018detectron}.
We fine-tune the model on AVA bounding boxes from the training videos for 130k iterations with an initial learning rate of 0.005,
which is decreased by a factor of 10 at iteration 100k and 120k. 
To improve generalization, we train with random scale jittering (from 512 to 800 pixels). 
The final model obtains 93.9 AP@50 on the AVA validation set.

%%%%%%%%%%%%%%%%%%%%%%%%%%%%%%%%%%%%%%%%%%%%%%%%%%%%%%%%%%%%%%%%%%%%%%%%%%%%%%%%%%%%%%%%%%%%%%%%%%%
\section{LFB \vs Improving Backbones}
A large body of recent research focuses on improving 3D CNN architectures, \ie, improving modeling of short-term patterns.
This paper, on the other hand, aims at improving the modeling of long-term patterns.
How do these two directions impact video understanding differently?

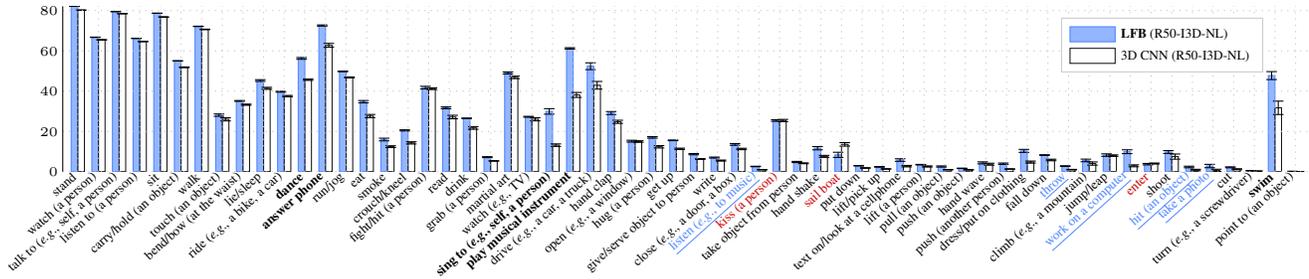
\begin{figure*}[h!]
  \tiny
  \center
  \begin{tikzpicture}
    \pgfplotstableread[col sep=semicolon]{%
category;
stand;
watch (a person);
talk to (\eg, self, a person);
listen to (a person);
sit;
carry/hold (an object);
walk;
touch (an object);
bend/bow (at the waist);
lie/sleep;
ride (\eg, a bike, a car);
\bf{dance};
\bf{answer phone};
run/jog;
eat;
smoke;
crouch/kneel;
fight/hit (a person);
read;
drink;
grab (a person);
martial art;
watch (\eg, TV);
\bf{sing to (\eg, self, a person)};
\bf{play musical instrument};
drive (\eg, a car, a truck);
hand clap;
open (\eg, a window);
hug (a person);
get up;
give/serve object to person;
write;
close (\eg, a door, a box);
{\color{fbblue1}\underline{listen (\eg, to music)}};
{\color{scarletred2}kiss (a person)};
take object from person;
hand shake;
{\color{scarletred2}sail boat};
put down;
lift/pick up;
text on/look at a cellphone;
lift (a person);
pull (an object);
push (an object);
hand wave;
push (another person);
dress/put on clothing;
fall down;
{\color{fbblue1}\underline{throw}};
climb (\eg, a mountain);
jump/leap;
{\color{fbblue1}\underline{work on a computer}};
{\color{scarletred2}enter};
shoot;
{\color{fbblue1}\underline{hit (an object)}};
{\color{fbblue1}\underline{take a photo}};
cut;
turn (\eg, a screwdriver);
\bf{swim};
point to (an object);
    }\data

    \pgfplotsset{shifti/.style={mark=no markers, bar width=2.5pt, bar shift=1.25pt,black!90}}
    \pgfplotsset{shiftt/.style={mark=no markers, bar width=2.5pt, bar shift=-1.25pt,fbblue1,fill=fbblue1!60}}
    \pgfplotsset{err/.style={forget plot, draw=none}} 
    \pgfplotsset{errp/.style={err, black, error bars/.cd,y dir=both,  y explicit}}

    \begin{axis}[%
    /pgfplots/table/header=false,
    scale only axis,
    width=0.95\textwidth,
    height=2.2cm,
    enlarge x limits={abs=0.2cm},
    xtick={1,...,60},
    xticklabels from table={\data}{category},
    x tick label style={rotate=42,anchor=east},
    axis on top,
    axis y line*=left,
    x axis line style={draw opacity=0},
    legend style={at={(0.8,0.6)},anchor=south west},
    area legend,
    ]
    \addplot+[ybar, shiftt] table[x index=0, y index=1]{plotdata/per_class_lfb_reverse.data};
    \addlegendentry{\tiny \textbf{LFB} (R50-I3D-NL)}
    \addplot+[no markers, xshift=-1.25pt, errp] table[x index=0, y index=1, y error index=2]{plotdata/per_class_lfb_reverse.data};

    \addplot+[ybar, shifti] table[x index=0, y index=1]{plotdata/per_class_baseline_reverse.data};
    \addlegendentry{\tiny 3D CNN (R50-I3D-NL)}
    \addplot+[no markers, xshift=1.25pt, errp] table[x index=0, y index=1, y error index=2]{plotdata/per_class_baseline_reverse.data};
    \end{axis}
  \end{tikzpicture}
% \vspace{-3mm}
\caption{\textbf{Impact of LFB.} We compare per-class AP of 3D CNN (22.1 mAP) and LFB model (25.5 mAP) on AVA\@.
LFB leads to larger improvement on \emph{interactive} actions, \eg, `sing to', `play musical instrument', or `work on a computer'.
(\textbf{Bold:} 5 classes with the largest absolute gain.
{\color{fbblue1}\underline{Blue:}} 5 classes with the largest relative gain.
{\color{scarletred2}Red: } classes with decreased performance.
Classes are sorted by frequency. )
\vspace{2mm}
}\label{fig:per_class_lfb}
\end{figure*}

\begin{figure*}[h!]
  \tiny
  \center
  \begin{tikzpicture}
    \pgfplotstableread[col sep=semicolon]{%
category;
stand;
watch (a person);
talk to (\eg, self, a person);
listen to (a person);
sit;
carry/hold (an object);
walk;
touch (an object);
bend/bow (at the waist);
lie/sleep;
{\color{scarletred2}ride (\eg, a bike, a car)};
dance;
answer phone;
run/jog;
{\color{scarletred2}eat};
smoke;
\bf{crouch/kneel};
fight/hit (a person);
\bf{read};
{\color{scarletred2}drink};
grab (a person);
martial art;
watch (\eg, TV);
\bf{sing to (\eg, self, a person)};
{\color{scarletred2}play musical instrument};
{\color{scarletred2}drive (\eg, a car, a truck)};
hand clap;
open (\eg, a window);
hug (a person);
get up;
give/serve object to person;
{\color{scarletred2}write};
close (\eg, a door, a box);
listen (\eg, to music);
kiss (a person);
take object from person;
{\color{fbblue1}\underline{\bf{hand shake}}};
sail boat;
put down;
lift/pick up;
{\color{scarletred2}text on/look at a cellphone};
lift (a person);
{\color{scarletred2}pull (an object)};
{\color{fbblue1}\underline{push (an object)}};
hand wave;
push (another person);
dress/put on clothing;
fall down;
{\color{fbblue1}\underline{throw}};
{\color{fbblue1}\underline{\bf{climb (\eg, a mountain)}}};
jump/leap;
work on a computer;
{\color{scarletred2}enter};
{\color{scarletred2}shoot};
{\color{scarletred2}hit (an object)};
{\color{fbblue1}\underline{take a photo}};
cut;
turn (\eg, a screwdriver);
{\color{scarletred2}swim};
point to (an object);
    }\data

    \pgfplotsset{shifti/.style={mark=no markers, bar width=2.5pt, bar shift=1.25pt,black!90}}
    \pgfplotsset{shiftt/.style={mark=no markers, bar width=2.5pt, bar shift=-1.25pt,fbred2,fill=fbred2!60}}
    \pgfplotsset{err/.style={forget plot, draw=none}} 
    \pgfplotsset{errp/.style={err, black, error bars/.cd,y dir=both,  y explicit}}

    \begin{axis}[%
    /pgfplots/table/header=false,
    scale only axis,
    width=0.95\textwidth,
    height=2.2cm,
    enlarge x limits={abs=0.2cm},
    xtick={1,...,60},
    xticklabels from table={\data}{category},
    x tick label style={rotate=42,anchor=east},
    axis on top,
    axis y line*=left,
    x axis line style={draw opacity=0},
    legend style={at={(0.8,0.6)},anchor=south west},
    area legend,
    ]
    \addplot+[ybar, shiftt] table[x index=0, y index=1]{plotdata/per_class_r101_reverse.data};
    \addlegendentry{\tiny 3D CNN \textbf{(R101-I3D-NL)}}
    \addplot+[no markers, xshift=-1.25pt, errp] table[x index=0, y index=1, y error index=2]{plotdata/per_class_r101_reverse.data};

    \addplot+[ybar, shifti] table[x index=0, y index=1]{plotdata/per_class_baseline_reverse.data};
    \addlegendentry{\tiny 3D CNN (R50-I3D-NL)}
    \addplot+[no markers, xshift=1.25pt, errp] table[x index=0, y index=1, y error index=2]{plotdata/per_class_baseline_reverse.data};
    \end{axis}
  \end{tikzpicture}
% \vspace{-3mm}
\caption{\textbf{Impact of improving backbone.} We compare per-class AP of 3D CNN with the default backbone (R50-I3D-NL; 22.1 mAP) and a stronger backbone (R101-I3D-NL; 23.0 mAP) on AVA\@.
Improving backbone leads to larger improvement in standalone actions, such as `crouch/kneel', `read', or `hand shake'.
(\textbf{Bold:} 5 classes with the largest absolute gain.
{\color{fbblue1}\underline{Blue:}} 5 classes with the largest relative gain.
{\color{scarletred2}Red: } classes with decreased performance.
)
\vspace{2mm}
}\label{fig:per_class_r101}
\end{figure*}
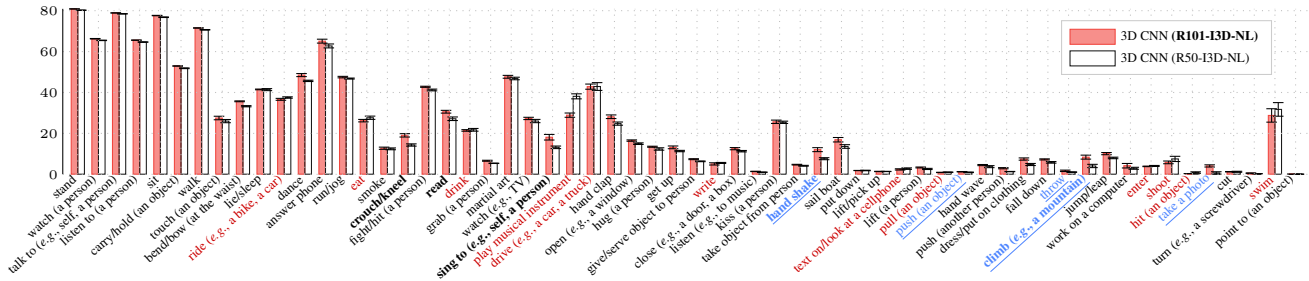

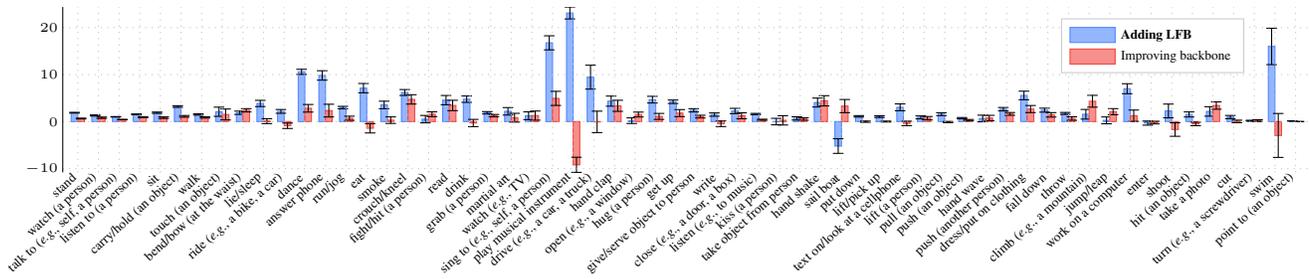
\begin{figure*}[h!]
  \tiny
  \center
  \begin{tikzpicture}
    \pgfplotstableread[col sep=semicolon]{%
category;
stand;
watch (a person);
talk to (\eg, self, a person);
listen to (a person);
sit;
carry/hold (an object);
walk;
touch (an object);
bend/bow (at the waist);
lie/sleep;
ride (\eg, a bike, a car);
dance;
answer phone;
run/jog;
eat;
smoke;
crouch/kneel;
fight/hit (a person);
read;
drink;
grab (a person);
martial art;
watch (\eg, TV);
sing to (\eg, self, a person);
play musical instrument;
drive (\eg, a car, a truck);
hand clap;
open (\eg, a window);
hug (a person);
get up;
give/serve object to person;
write;
close (\eg, a door, a box);
listen (\eg, to music);
kiss (a person);
take object from person;
hand shake;
sail boat;
put down;
lift/pick up;
text on/look at a cellphone;
lift (a person);
pull (an object);
push (an object);
hand wave;
push (another person);
dress/put on clothing;
fall down;
throw;
climb (\eg, a mountain);
jump/leap;
work on a computer;
enter;
shoot;
hit (an object);
take a photo;
cut;
turn (\eg, a screwdriver);
swim;
point to (an object);
    }\data

    \pgfplotsset{shifti/.style={mark=no markers, bar width=2.5pt, bar shift=1.25pt,fbred2,fill=fbred2!60}}
    \pgfplotsset{shiftt/.style={mark=no markers, bar width=2.5pt, bar shift=-1.25pt,fbblue1,fill=fbblue1!60}}
    \pgfplotsset{err/.style={forget plot, draw=none}} 
    \pgfplotsset{errp/.style={err, black, error bars/.cd,y dir=both,  y explicit}}

    \begin{axis}[%
    /pgfplots/table/header=false,
    scale only axis,
    width=0.95\textwidth,
    height=2.2cm,
    enlarge x limits={abs=0.2cm},
    xtick={1,...,60},
    xticklabels from table={\data}{category},
    x tick label style={rotate=42,anchor=east},
    axis on top,
    axis y line*=left,
    x axis line style={draw opacity=0},
    legend style={at={(0.8,0.6)},anchor=south west},
    area legend,
    ]
    \addplot+[ybar, shiftt] table[x index=0, y index=1]{plotdata/per_class_lfb_gain_reverse.data};
    \addlegendentry{\tiny \bf{Adding LFB}}
    \addplot+[no markers, xshift=-1.25pt, errp] table[x index=0, y index=1, y error index=2]{plotdata/per_class_lfb_gain_reverse.data};

    \addplot+[ybar, shifti] table[x index=0, y index=1]{plotdata/per_class_r101_gain_reverse.data};
    \addlegendentry{\tiny Improving backbone}
    \addplot+[no markers, xshift=1.25pt, errp] table[x index=0, y index=1, y error index=2]{plotdata/per_class_r101_gain_reverse.data};
    \end{axis}
  \end{tikzpicture}
% \vspace{-3mm}
\caption{\textbf{Adding LFB \vs improving backbone.} 
We compare the absolute improvement (in AP) brought by LFB and the improvement brought by improving backbone. 
We see that they lead to improvement in different action classes.
This suggests that improving long-term modeling (through LFB) and short-term modeling (through improving backbone) are complementary; we believe
both are important for future video understanding research.\vspace{1mm}
}\label{fig:per_class_gain}
\end{figure*}

We plot the per-class impact of LFB in \figref{per_class_lfb},
the per-class impact of improving backbone in \figref{per_class_r101},
and compare them in \figref{per_class_gain}.
The error bars are plus/minus one standard error around the mean, computed from 5 runs.
We see that they lead to improvement in different action classes.
Using LFB leads to improvement in many interactive actions, such as `play musical instrument' or `sing to',
while improving backbone leads to improvement in more standalone actions, such as `hand shake' or `climb'.
This suggests that improving long-term modeling (through LFB) and short-term modeling (through improving backbone) are complementary; we believe
both are important for future video understanding research.

%##################################################################################################
%##################################################################################################

%##################################################################################################
%##################################################################################################

%##################################################################################################
%##################################################################################################

%%%%%%%%%%%%%%%%%%%%%%%%%%%%%%%%%%%%%%%%%%%%%%%%%%%%%%%%%%%%%%%%%%%%%%%%%%%%%%%%%%%%%%%%%%%%%%%%%%%
\section{AVA STO Regularization}
To address the overfitting issue of $\stm$ on AVA, 
we experimented with a number of regularization techniques, as summarized in \tabref{sto}.
We found that dropout~\cite{srivastava2014dropout} was insufficient to regularize an $\stm$,
but injecting `distractors', \ie, randomly sampled features, into the features being attended during training was very effective.
We report $\stm$ results with `distractor' training for AVA unless otherwise stated.
For $\stm$ on other datasets, we did not observe obvious overfitting. 
\begin{table}[h]
\tablestyle{7.9pt}{1.0}\begin{tabular}{@{}ccccc@{}}
3D CNN & $\stm$ & $\stm$ & $\stm$ & $\stm$\\
& & + 0.5 dropout & + 0.8 dropout & + distractors\\
\shline
22.1 & 20.2 & 20.1 & 20.9 & \bf 23.2
\end{tabular}
\vspace{1mm}
\caption{\textbf{$\stm$ with different regularization techniques on AVA} (mAP in \%).}\label{tab:sto}
\end{table}

%%%%%%%%%%%%%%%%%%%%%%%%%%%%%%%%%%%%%%%%%%%%%%%%%%%%%%%%%%%%%%%%%%%%%%%%%%%%%%%%%%%%%%%%%%%%%%%%%%%
\section{Training Schedule for EPIC-Kitchens}
We train the verb models for 36k iterations with $10^{-5}$ weight decay and a learning rate of 0.0003,
which is then decreased by 10 times at iteration 28k and 32k. 
For the noun models, we train for 50k iterations with weight decay $10^{-6}$ and a learning rate of 0.001,
which is decreased by 10 times at iteration 40k and 45k. 

%%%%%%%%%%%%%%%%%%%%%%%%%%%%%%%%%%%%%%%%%%%%%%%%%%%%%%%%%%%%%%%%%%%%%%%%%%%%%%%%%%%%%%%%%%%%%%%%%%%
\section{EPIC-Kitchens Inference}
We sample training clips such that the center of the clip falls within a training segment.
For testing, we sample one center clip per segment, resize such that the short side is 256 pixels, and use a single center crop of 256$\x$256.
We compute the probability of an action as the product of the softmax scores, weighted by a prior $\mu$, \ie
\begin{align*}
P(\mathrm{action}=(v, n)) \propto \mu(v, n) P(\mathrm{verb}=v) P(\mathrm{noun}=n),
\end{align*}
where $\mu$ is a prior estimated as the count of $(v, n)$ pair divided by count of $n$ in training annotations.

%%%%%%%%%%%%%%%%%%%%%%%%%%%%%%%%%%%%%%%%%%%%%%%%%%%%%%%%%%%%%%%%%%%%%%%%%%%%%%%%%%%%%%%%%%%%%%%%%%%
\section{Object Detector for LFB of EPIC-Kitchens Noun Model}
We use Faster R-CNN~\cite{Ren2015a} with ResNeXt-101-FPN~\cite{xie2017aggregated,lin2017feature} backbone for the object detector. 
The detector is pre-trained on Visual Genome~\cite{krishna2017visual} with 1600 class labels defined in Anderson~\etal~\cite{anderson2018bottom}.
We fine-tune this model on the `new' training split (defined in Baradel~\etal~\cite{baradel2018object}) of EPIC-Kitchens
for 90k iterations, with random scale jittering (from 512 to 800 pixels).
We use an initial learning rate of 0.005, which is decreased by a factor of 10 at iteration 60k and 80k.
The final model achieves 2.4 AP on the `new' validation split.
The AP is low because with the new train/val split, most of the classes are unseen during training.
In addition, most of the classes have zero instance in the new, smaller validation set, and we calculate the average precision of those classes as 0. 
This is also not comparable to the performance reported in \cite{epic}, where the model is trained on the full training set, and evaluated on the unreleased test sets.

%%%%%%%%%%%%%%%%%%%%%%%%%%%%%%%%%%%%%%%%%%%%%%%%%%%%%%%%%%%%%%%%%%%%%%%%%%%%%%%%%%%%%%%%%%%%%%%%%%%
\section{Charades Training Schedule}
We train the models to predict the `clip-level' labels,
\ie, the union of the frame labels that fall into the clip's temporal range.
We train the baseline 3D CNNs with the default 24k schedule with learning rate 0.02,
which is decreased by a factor of 10 at iteration 20k.
To train LFB models, we use a 2-stage approach following STRG~\cite{strg}.
We first train the model without the FBO using the 24k schedule, and then add FBO, 
freeze backbone, and train for half of the schedule (12k iterations, so 36k in total).
This schedule helps prevent overfitting that was observed when training directly with FBO in one stage.
For training STO, we observed worse performance with this 2-stage approach, 
so we report the STO performance using the default 24k schedule.

%%%%%%%%%%%%%%%%%%%%%%%%%%%%%%%%%%%%%%%%%%%%%%%%%%%%%%%%%%%%%%%%%%%%%%%%%%%%%%%%%%%%%%%%%%%%%%%%%%%
\section{Charades NL Block Details}

%##################################################################################################
\begin{table}[h]
\tablestyle{7pt}{1.0}\begin{tabular}{@{}l|cc|cc@{}}
  &\multicolumn{2}{c|}{R50-I3D-NL} & \multicolumn{2}{c}{R101-I3D-NL}\\
  & \demph{pre-act} & \bf post-act & \demph{pre-act} & \bf post-act\\
\shline
  {STO}         & \demph{39.0} & 39.6 & \demph{40.5} & 41.0 \\
\textbf{LFB NL} & \demph{39.5} & \bf40.3  & \demph{41.5} & \bf42.5\\
\hline
  {3D CNN}      & \multicolumn{2}{c|}{38.3} & \multicolumn{2}{c}{40.3}\\
\end{tabular}
\vspace{1mm}
\caption{\textbf{Pre-activation \vs post-activation $\nlmod$ on Charades.} }\label{tab:charades_variants}
\end{table}
%##################################################################################################

We experimented with two variants of $\nlmod$ design:
a \emph{pre-activation} (the default variant described in paper), 
and a \emph{post-activation} variant, where we move ReLU after the skip connection, 
and move the layer normalization after the output linear layer. 
For both variants, LFB consistently outperforms STO and baseline 3D CNN (\tabref{charades_variants}).
We choose post-activation as default for Charades due to the stronger performance.
For AVA and EPIC-Kitchens, we did not observe any noticeable difference between the two variants.

\end{appendices}

{\small
\bibliographystyle{ieee}
\bibliography{lfb}

\begin{thebibliography}{10}\itemsep=-1pt

\bibitem{althoff2012detection}
T.~Althoff, H.~O. Song, and T.~Darrell.
\newblock Detection bank: an object detection based video representation for
  multimedia event recognition.
\newblock In {\em International Conference on Multimedia}, 2012.

\bibitem{anderson2018bottom}
P.~Anderson, X.~He, C.~Buehler, D.~Teney, M.~Johnson, S.~Gould, and L.~Zhang.
\newblock Bottom-up and top-down attention for image captioning and visual
  question answering.
\newblock In {\em CVPR}, 2018.

\bibitem{ba2016layer}
J.~L. Ba, J.~R. Kiros, and G.~E. Hinton.
\newblock Layer normalization.
\newblock {\em arXiv preprint arXiv:1607.06450}, 2016.

\bibitem{baradel2018object}
F.~Baradel, N.~Neverova, C.~Wolf, J.~Mille, and G.~Mori.
\newblock Object level visual reasoning in videos.
\newblock In {\em ECCV}, 2018.

\bibitem{i3d}
J.~Carreira and A.~Zisserman.
\newblock Quo vadis, action recognition? a new model and the kinetics dataset.
\newblock In {\em CVPR}, 2017.

\bibitem{epic}
D.~Damen, H.~Doughty, G.~M. Farinella, S.~Fidler, A.~Furnari, E.~Kazakos,
  D.~Moltisanti, J.~Munro, T.~Perrett, W.~Price, et~al.
\newblock Scaling egocentric vision: The {EPIC}-kitchens dataset.
\newblock In {\em ECCV}, 2018.

\bibitem{donahue2015long}
J.~Donahue, L.~Anne~Hendricks, S.~Guadarrama, M.~Rohrbach, S.~Venugopalan,
  K.~Saenko, and T.~Darrell.
\newblock Long-term recurrent convolutional networks for visual recognition and
  description.
\newblock In {\em CVPR}, 2015.

\bibitem{feichtenhofer2016spatiotemporal}
C.~Feichtenhofer, A.~Pinz, and R.~Wildes.
\newblock Spatiotemporal residual networks for video action recognition.
\newblock In {\em NIPS}, 2016.

\bibitem{girdhar2018better}
R.~Girdhar, J.~Carreira, C.~Doersch, and A.~Zisserman.
\newblock A better baseline for {AVA}.
\newblock {\em arXiv preprint arXiv:1807.10066}, 2018.

\bibitem{girshick2015fast}
R.~Girshick.
\newblock Fast {R-CNN}.
\newblock In {\em ICCV}, 2015.

\bibitem{girshick2018detectron}
R.~Girshick, I.~Radosavovic, G.~Gkioxari, P.~Doll{\'a}r, and K.~He.
\newblock Detectron, 2018.

\bibitem{gkioxari2015finding}
G.~Gkioxari and J.~Malik.
\newblock Finding action tubes.
\newblock In {\em CVPR}, 2015.

\bibitem{goyal2017imagenet1hr}
P.~Goyal, P.~Doll\'{a}r, R.~Girshick, P.~Noordhuis, L.~Wesolowski, A.~Kyrola,
  A.~Tulloch, Y.~Jia, and K.~He.
\newblock Accurate, large minibatch sgd: Training imagenet in 1 hour.
\newblock {\em arXiv preprint arXiv:1706.02677}, 2017.

\bibitem{ava}
C.~Gu, C.~Sun, D.~A. Ross, C.~Vondrick, C.~Pantofaru, Y.~Li,
  S.~Vijayanarasimhan, G.~Toderici, S.~Ricco, R.~Sukthankar, et~al.
\newblock {AVA}: A video dataset of spatio-temporally localized atomic visual
  actions.
\newblock In {\em CVPR}, 2018.

\bibitem{He2017}
K.~He, G.~Gkioxari, P.~Doll{\'a}r, and R.~Girshick.
\newblock Mask {R-CNN}.
\newblock In {\em ICCV}, 2017.

\bibitem{He2016}
K.~He, X.~Zhang, S.~Ren, and J.~Sun.
\newblock Deep residual learning for image recognition.
\newblock In {\em CVPR}, 2016.

\bibitem{hou2017tube}
R.~Hou, C.~Chen, and M.~Shah.
\newblock Tube convolutional neural network {(T-CNN)} for action detection in
  videos.
\newblock In {\em ICCV}, 2017.

\bibitem{ioffe2015batch}
S.~Ioffe and C.~Szegedy.
\newblock Batch normalization: Accelerating deep network training by reducing
  internal covariate shift.
\newblock In {\em ICML}, 2015.

\bibitem{jianghuman}
J.~Jiang, Y.~Cao, L.~Song, S.~Z.~Y. Li, Z.~Xu, Q.~Wu, C.~Gan, C.~Zhang, and
  G.~Yu.
\newblock Human centric spatio-temporal action localization.
\newblock In {\em ActivityNet workshop, CVPR}, 2018.

\bibitem{kalogeiton2017action}
V.~Kalogeiton, P.~Weinzaepfel, V.~Ferrari, and C.~Schmid.
\newblock Action tubelet detector for spatio-temporal action localization.
\newblock In {\em ICCV}, 2017.

\bibitem{karpathy2014large}
A.~Karpathy, G.~Toderici, S.~Shetty, T.~Leung, R.~Sukthankar, and L.~Fei-Fei.
\newblock Large-scale video classification with convolutional neural networks.
\newblock In {\em CVPR}, 2014.

\bibitem{krishna2017visual}
R.~Krishna, Y.~Zhu, O.~Groth, J.~Johnson, K.~Hata, J.~Kravitz, S.~Chen,
  Y.~Kalantidis, L.-J. Li, D.~A. Shamma, et~al.
\newblock Visual genome: Connecting language and vision using crowdsourced
  dense image annotations.
\newblock {\em IJCV}, 2017.

\bibitem{lecun1986learning}
Y.~LeCun.
\newblock Learning process in an asymmetric threshold network.
\newblock In {\em Disordered systems and biological organization}. 1986.

\bibitem{li2018recurrent}
D.~Li, Z.~Qiu, Q.~Dai, T.~Yao, and T.~Mei.
\newblock Recurrent tubelet proposal and recognition networks for action
  detection.
\newblock In {\em ECCV}, 2018.

\bibitem{li2017temporal}
F.~Li, C.~Gan, X.~Liu, Y.~Bian, X.~Long, Y.~Li, Z.~Li, J.~Zhou, and S.~Wen.
\newblock Temporal modeling approaches for large-scale youtube-8m video
  understanding.
\newblock {\em arXiv preprint arXiv:1707.04555}, 2017.

\bibitem{li2010object}
L.-J. Li, H.~Su, L.~Fei-Fei, and E.~P. Xing.
\newblock Object bank: A high-level image representation for scene
  classification \& semantic feature sparsification.
\newblock In {\em NIPS}, 2010.

\bibitem{li2018videolstm}
Z.~Li, K.~Gavrilyuk, E.~Gavves, M.~Jain, and C.~G. Snoek.
\newblock Videolstm convolves, attends and flows for action recognition.
\newblock {\em Computer Vision and Image Understanding}, 2018.

\bibitem{lin2017feature}
T.-Y. Lin, P.~Doll{\'a}r, R.~B. Girshick, K.~He, B.~Hariharan, and S.~J.
  Belongie.
\newblock Feature pyramid networks for object detection.
\newblock In {\em CVPR}, 2017.

\bibitem{lin2014microsoft}
T.-Y. Lin, M.~Maire, S.~Belongie, J.~Hays, P.~Perona, D.~Ramanan,
  P.~Doll{\'a}r, and C.~L. Zitnick.
\newblock Microsoft coco: Common objects in context.
\newblock In {\em ECCV}, 2014.

\bibitem{ma2017attend}
C.-Y. Ma, A.~Kadav, I.~Melvin, Z.~Kira, G.~AlRegib, and H.~P. Graf.
\newblock Attend and interact: Higher-order object interactions for video
  understanding.
\newblock In {\em CVPR}, 2018.

\bibitem{miech2017learnable}
A.~Miech, I.~Laptev, and J.~Sivic.
\newblock Learnable pooling with context gating for video classification.
\newblock {\em arXiv preprint arXiv:1706.06905}, 2017.

\bibitem{peng2016multi}
X.~Peng and C.~Schmid.
\newblock Multi-region two-stream r-cnn for action detection.
\newblock In {\em ECCV}, 2016.

\bibitem{qiu2017learning}
Z.~Qiu, T.~Yao, and T.~Mei.
\newblock Learning spatio-temporal representation with pseudo-3d residual
  networks.
\newblock In {\em ICCV}, 2017.

\bibitem{Ren2015a}
S.~Ren, K.~He, R.~Girshick, and J.~Sun.
\newblock {Faster R-CNN}: Towards real-time object detection with region
  proposal networks.
\newblock In {\em NIPS}, 2015.

\bibitem{Russakovsky2015}
O.~Russakovsky, J.~Deng, H.~Su, J.~Krause, S.~Satheesh, S.~Ma, Z.~Huang,
  A.~Karpathy, A.~Khosla, M.~Bernstein, A.~C. Berg, and L.~Fei-Fei.
\newblock {ImageNet Large Scale Visual Recognition Challenge}.
\newblock {\em IJCV}, 2015.

\bibitem{saha2017amtnet}
S.~Saha, G.~Singh, and F.~Cuzzolin.
\newblock {AMT}net: Action-micro-tube regression by end-to-end trainable deep
  architecture.
\newblock In {\em ICCV}, 2017.

\bibitem{sigurdsson2017asynchronous}
G.~A. Sigurdsson, S.~K. Divvala, A.~Farhadi, and A.~Gupta.
\newblock Asynchronous temporal fields for action recognition.
\newblock In {\em CVPR}, 2017.

\bibitem{charades}
G.~A. Sigurdsson, G.~Varol, X.~Wang, A.~Farhadi, I.~Laptev, and A.~Gupta.
\newblock Hollywood in homes: Crowdsourcing data collection for activity
  understanding.
\newblock In {\em ECCV}, 2016.

\bibitem{simonyan2014two}
K.~Simonyan and A.~Zisserman.
\newblock Two-stream convolutional networks for action recognition in videos.
\newblock In {\em NIPS}, 2014.

\bibitem{singh2017online}
G.~Singh, S.~Saha, M.~Sapienza, P.~H. Torr, and F.~Cuzzolin.
\newblock Online real-time multiple spatiotemporal action localisation and
  prediction.
\newblock In {\em ICCV}, 2017.

\bibitem{srivastava2014dropout}
N.~Srivastava, G.~Hinton, A.~Krizhevsky, I.~Sutskever, and R.~Salakhutdinov.
\newblock Dropout: a simple way to prevent neural networks from overfitting.
\newblock {\em JMLR}, 2014.

\bibitem{sukhbaatar2015end}
S.~Sukhbaatar, A.~Szlam, J.~Weston, and R.~Fergus.
\newblock End-to-end memory networks.
\newblock In {\em NIPS}, 2015.

\bibitem{sun2018actor}
C.~Sun, A.~Shrivastava, C.~Vondrick, K.~Murphy, R.~Sukthankar, and C.~Schmid.
\newblock Actor-centric relation network.
\newblock In {\em ECCV}, 2018.

\bibitem{sun2017lattice}
L.~Sun, K.~Jia, K.~Chen, D.-Y. Yeung, B.~E. Shi, and S.~Savarese.
\newblock Lattice long short-term memory for human action recognition.
\newblock In {\em ICCV}, 2017.

\bibitem{tang2018non}
Y.~Tang, X.~Zhang, J.~Wang, S.~Chen, L.~Ma, and Y.-G. Jiang.
\newblock Non-local netvlad encoding for video classification.
\newblock {\em arXiv preprint arXiv:1810.00207}, 2018.

\bibitem{tran2015learning}
D.~Tran, L.~Bourdev, R.~Fergus, L.~Torresani, and M.~Paluri.
\newblock Learning spatiotemporal features with 3d convolutional networks.
\newblock In {\em ICCV}, 2015.

\bibitem{tran2017convnet}
D.~Tran, J.~Ray, Z.~Shou, S.-F. Chang, and M.~Paluri.
\newblock Convnet architecture search for spatiotemporal feature learning.
\newblock {\em arXiv preprint arXiv:1708.05038}, 2017.

\bibitem{varol2018long}
G.~Varol, I.~Laptev, and C.~Schmid.
\newblock Long-term temporal convolutions for action recognition.
\newblock {\em PAMI}, 2018.

\bibitem{vaswani2017attention}
A.~Vaswani, N.~Shazeer, N.~Parmar, J.~Uszkoreit, L.~Jones, A.~N. Gomez,
  {\L}.~Kaiser, and I.~Polosukhin.
\newblock Attention is all you need.
\newblock In {\em NIPS}, 2017.

\bibitem{tsn}
L.~Wang, Y.~Xiong, Z.~Wang, Y.~Qiao, D.~Lin, X.~Tang, and L.~Van~Gool.
\newblock Temporal segment networks: Towards good practices for deep action
  recognition.
\newblock In {\em ECCV}, 2016.

\bibitem{nonlocal}
X.~Wang, R.~Girshick, A.~Gupta, and K.~He.
\newblock Non-local neural networks.
\newblock In {\em CVPR}, 2018.

\bibitem{strg}
X.~Wang and A.~Gupta.
\newblock Videos as space-time region graphs.
\newblock In {\em ECCV}, 2018.

\bibitem{weinzaepfel2015learning}
P.~Weinzaepfel, Z.~Harchaoui, and C.~Schmid.
\newblock Learning to track for spatio-temporal action localization.
\newblock In {\em ICCV}, 2015.

\bibitem{coviar}
C.-Y. Wu, M.~Zaheer, H.~Hu, R.~Manmatha, A.~J. Smola, and
  P.~Kr{\"a}henb{\"u}hl.
\newblock Compressed video action recognition.
\newblock In {\em CVPR}, 2018.

\bibitem{xie2017aggregated}
S.~Xie, R.~Girshick, P.~Doll{\'a}r, Z.~Tu, and K.~He.
\newblock Aggregated residual transformations for deep neural networks.
\newblock In {\em CVPR}, 2017.

\bibitem{xie2017rethinking}
S.~Xie, C.~Sun, J.~Huang, Z.~Tu, and K.~Murphy.
\newblock Rethinking spatiotemporal feature learning for video understanding.
\newblock In {\em ECCV}, 2018.

\bibitem{yue2015beyond}
J.~Yue-Hei~Ng, M.~Hausknecht, S.~Vijayanarasimhan, O.~Vinyals, R.~Monga, and
  G.~Toderici.
\newblock Beyond short snippets: Deep networks for video classification.
\newblock In {\em CVPR}, 2015.

\bibitem{zhou2017temporalrelation}
B.~Zhou, A.~Andonian, A.~Oliva, and A.~Torralba.
\newblock Temporal relational reasoning in videos.
\newblock In {\em ECCV}, 2018.

\end{thebibliography}
}

\end{document}